\documentclass[a4paper,fleqn]{cas-sc}
\usepackage[numbers]{natbib}
\usepackage{amssymb}
\usepackage{amsmath}
\usepackage{booktabs}
\usepackage{graphicx}
\usepackage{caption}
\usepackage{subcaption}
\usepackage{siunitx}
\usepackage{xcolor}
\usepackage{placeins}
\usepackage{gensymb}
\usepackage{makecell}    
\usepackage{adjustbox}   

\usepackage{hyperref}
\hypersetup{hidelinks,
	colorlinks=true,
	allcolors=blue,
	pdfstartview=Fit,
	breaklinks=true}

\usepackage{amsmath}
\usepackage{amssymb}
\usepackage{multirow}
\usepackage{adjustbox}
\usepackage{makecell}

\usepackage{lineno}

\def\tsc#1{\csdef{#1}{\textsc{\lowercase{#1}}\xspace}}
\tsc{WGM}
\tsc{QE}
\tsc{EP}
\tsc{PMS}
\tsc{BEC}
\tsc{DE}

\begin{document}
\let\WriteBookmarks\relax
\def\floatpagepagefraction{1}
\def\textpagefraction{.001}
\newcommand{\code}[1]{\texttt{#1}}
\shorttitle{STARS}

\shortauthors{Tong Wang et~al.}

\title [mode = title]{STARS: Shared-specific Translation and Alignment for missing-modality Remote Sensing Semantic Segmentation}                      



%
\author[1]{Tong Wang}[type=editor,
                        orcid=0000-0002-3915-0175]

\ead{kingcopper@whu.edu.cn}
\credit{Conceptualization of this study, Methodology, Software, Data curation, Writing - Original draft preparation}

\author[1]{Xiaodong Zhang}[style=chinese]
\cormark[1]
\ead{zxdlmars@whu.edu.cn}
\credit{Supervision, Project administration, Funding acquisition}

\author[1]{Guanzhou Chen}[style=chinese]
\cormark[1]
\ead{cgz@whu.edu.cn}
\credit{Data curation, Writing - Original draft preparation}

\author[1]{Jiaqi Wang}
\credit{Writing - Review \& Editing, Resources}

\author[2]{Chenxi Liu}
\credit{Writing - Review \& Editing, Visualization}

\author[1]{Xiaoliang Tan}
\credit{Writing - Review \& Editing}

\author[1]{Wenchao Guo}
\credit{Writing - Review \& Editing}

\author[1]{Xuyang Li}
\credit{Writing - Review \& Editing}

\author[1]{Xuanrui Wang}
\credit{Writing - Review \& Editing}

\author[1,3]{Zifan Wang}
\credit{Writing - Review \& Editing}

\affiliation[1]{organization={State Key Laboratory of information Engineering in Surveying, Mapping and Remote Sensing, Wuhan University},
    addressline={299 Bayi Road, Wuchang District}, 
    city={Wuhan},
    postcode={430079}, 
    country={China}}

\affiliation[2]{organization={Electronic Information School, Wuhan University},
    addressline={299 Bayi Road, Wuchang District}, 
    city={Wuhan},
    postcode={430072}, 
    country={China}}

\affiliation[3]{organization={Hubei FreerTech Co. Ltd},
    addressline={No.2 Wenhua Road, Guandong Industrial Park, East Lake High-tech Development Zone}, 
    city={Wuhan},
    postcode={430073}, 
    state={Hubei},
    country={China}
}



\cortext[cor1]{Corresponding authors: zxdlmars@whu.edu.cn (X. Zhang), cgz@whu.edu.cn (G. Chen).}

\begin{abstract}
Multimodal remote sensing technology significantly enhances the understanding of surface semantics by integrating heterogeneous data such as optical images, Synthetic Aperture Radar (SAR), and Digital Surface Models (DSM). However, in practical applications, the missing of modality data (e.g., optical or DSM) is a common and severe challenge, which leads to performance decline in traditional multimodal fusion models. Existing methods for addressing missing modalities still face limitations, including feature collapse and overly generalized recovered features. 
To address these issues, we propose \textbf{STARS} (\textbf{S}hared-specific \textbf{T}ranslation and \textbf{A}lignment for missing-modality \textbf{R}emote \textbf{S}ensing), a robust semantic segmentation framework for incomplete multimodal inputs. STARS is built on two key designs. First, we introduce an asymmetric alignment mechanism with bidirectional translation and stop-gradient, which effectively prevents feature collapse and reduces sensitivity to hyperparameters. Second, we propose a Pixel-level Semantic sampling Alignment (PSA) strategy that combines class-balanced pixel sampling with cross-modality semantic alignment loss, to mitigate alignment failures caused by severe class imbalance and improve minority-class recognition.
Experiments on EarthMiss, WHU-OPT-SAR, and ISPRS Potsdam demonstrate that STARS consistently outperforms state-of-the-art methods under missing-modality settings. In particular, STARS achieves 33.04 $mIoU$  (+9.60 vs. Baseline-SAR) on EarthMiss (SAR-only test), and reaches 34.07 $mIoU$ (+5.95 vs. Baseline-SAR) on WHU-OPT-SAR under the same missing-modality evaluation. Additional analyses further verify that STARS establishes reliable cross-modality semantic correspondence while decoupling shared semantics from modality-specific cues, enabling accurate and stable remote sensing semantic segmentation with incomplete multimodal data.

\end{abstract}








\begin{keywords}
Multimodal Data Fusion \sep Remote Sensing \sep Semantic Segmentation \sep Digital Surface Model
\end{keywords}

\maketitle


\section{Introduction}
\label{sec:intro}
Multimodal remote sensing data, such as optical images, Synthetic Aperture Radar (SAR) images, and Digital Surface Models (DSM), greatly enrich our understanding of the Earth's surface~\cite{wang2025lmfnet, aleissaee2023transformers, he2023mftransnet,wang2026mssdf}. The fusion of multimodal remote sensing data has been widely applied to tasks such as land cover classification, object detection, and semantic segmentation~\cite{xie2025microwave,schmitt2017fusion,audebert2018beyond,li2022deep,XU202421,kang2022cfnet}. This is because it can leverage the complementary information from different modalities, thereby enhancing the overall performance and robustness of the models~\cite{he2023mftransnet, wang2025lmfnet, chen2024acmfnet,liu2025deep,liu2025estimation}. Specifically, remote sensing semantic segmentation aims to assign a semantic category to each pixel and serves as the foundation for critical applications such as land cover mapping, urban planning, disaster monitoring, and environmental management~\cite{chen2018symmetrical, chen2018training}. By integrating information from different sensors, the model can obtain a more comprehensive set of features. For example, optical images provide rich texture and spectral information, DSMs provide elevation and structural information, and SAR images can penetrate clouds to offer all-weather surface structure information~\cite{li2022deep,roy2023multimodal,adrian2021sentinel}.

However, in practical remote sensing applications, the integrity of multimodal data is a common and severe challenge~\cite{kang2022disoptnet,kampffmeyer2018urban,shermeyer2020spacenet}. Auxiliary modality data are often partially missing or completely unavailable. This can be due to sensor failures, environmental conditions (such as cloud cover, haze, nighttime imaging, terrain occlusion), or data acquisition cost constraints~\cite{ma2024multilevel,shermeyer2020spacenet,li2022deep,kampffmeyer2018urban}. The unreliability of this data makes traditional multimodal fusion models, which heavily rely on the complete availability of all modalities, vulnerable~\cite{zhou2026remote}. When auxiliary modalities are missing, it directly leads to blurred boundaries of ground objects and an increased rate of classification errors~\cite{kang2022disoptnet}.

\begin{figure}[htb]
    \centering
    \includegraphics[width=0.9\linewidth]{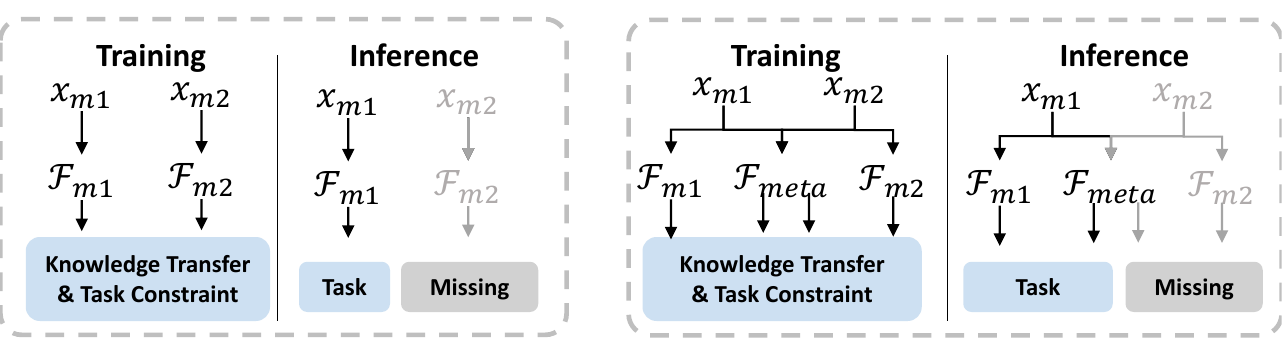}
    \caption{
    Comparison of existing methods for handling missing modalities. On the left, methods based on knowledge distillation or mutual learning are shown, which aim to transfer knowledge from complete modalities to the missing scenarios. On the right, methods based on shared-specific feature learning are displayed, which compensate for the missing modality by decoupling features and utilizing shared/meta features ($ \mathcal{F}_{meta}$).
    }
    \label{fig:prework}
\end{figure}

To address the challenge of modality missing, existing research has proposed various methods. These methods can be broadly categorized into two types, as shown in Figure~\ref{fig:prework}. One type focuses on knowledge distillation or modality mutual learning. These methods aim to transfer knowledge from complete modalities to scenarios with missing modalities~\cite{zheng2021deep,zhang2018deep,kang2022disoptnet,wang2021acn}. The other type of methods focuses on learning shared and specific features. They aim to enhance the model's adaptability to missing modalities by decoupling these features~\cite{zhou2026remote}.
For example, ShaSpec~\cite{wang2023multi} learns shared and specific features through auxiliary tasks such as distribution alignment and domain classification. When auxiliary modalities are missing, it uses the available shared features from the remaining modalities to compensate. MetaRS~\cite{zhou2026remote} introduces a meta-modality awareness module to extract modality-invariant features. It achieves feature decoupling through the supervision of covariance matrices.

However, existing methods still have some limitations when addressing the modality missing problem in remote sensing semantic segmentation. First, there are issues regarding feature collapse and hyperparameter sensitivity \cite{chen2021exploring, zhou2026remote}. When using a siamese architecture for cross-modality alignment, the model is prone to falling into a trivial solution. This means the encoder outputs the same constant vector for all inputs, causing the features to lose discriminative power. Traditional methods often rely on extremely strict loss weight adjustments (e.g., setting the alignment loss weight to a very small value) to maintain training stability \cite{wang2023multi}, which reduces the algorithm's robustness.

Second, significant class imbalance is common in remote sensing scenarios. Large-area background features, such as water bodies and forests, absolutely dominate in terms of pixel count, while the pixel proportion of key targets like buildings and roads is extremely small. Conventional global alignment strategies cause the model to overfit the features of majority classes. As a result, the cross-modality semantic alignment of minority key targets is marginalized or even completely fails during training \cite{zhou2026remote}.

To overcome the above challenges, we propose \textbf{STARS} (\textbf{S}hared-specific \textbf{T}ranslation and \textbf{A}lignment for multimodal \textbf{R}emote \textbf{S}ensing), a robust semantic segmentation framework for multimodal remote sensing data. The core of STARS is a shared-specific feature modeling architecture. It integrates two key techniques: (1) an asymmetric gradient control mechanism to achieve hyperparameter-robust alignment, and (2) a pixel-level semantic balanced sampling strategy for more discriminative cross-modality knowledge transfer. The performance advantages of STARS are mainly achieved through: an asymmetric alignment mechanism based on bidirectional translation and gradient stopping, and a pixel-level semantic sampling alignment strategy (PSA).



In summary, STARS provides a reliable framework for the application of multimodal remote sensing imagery in complex surface environments through its innovative asymmetric alignment and semantic balancing sampling strategies. The main contributions of this study are summarized as follows:

\begin{itemize}
    \item We propose STARS, a robust semantic segmentation framework for multimodal remote sensing data, which mitigates instability in cross-modality feature alignment through a Shared-Specific architecture.
    \item We reveal and address the hyperparameter sensitivity inherent in symmetric cross-modality alignment by proposing an asymmetric mechanism, making training more stable and practical.
    \item We design a Pixel-level Semantic Sampling Alignment (PSA) strategy. Through balanced sampling and cross-modality consistency loss, this strategy alleviates alignment failures caused by class imbalance in remote sensing images and improves the recognition accuracy of minority-class targets.
\end{itemize}

This paper is organized as follows: Section~\ref{sec:related} reviews related work; Section~\ref{sec:method} presents the architecture of STARS in detail. Section~\ref{sec:experiments} reports experimental results and ablation studies; Section~\ref{sec:disscussion} discusses how STARS establishes robust semantic bridges and decouples modality-specific features, and Section~\ref{sec:conclusion} concludes the paper with future directions.

\section{Related Work}
\label{sec:related}

\subsection{Knowledge Distillation and modality Mutual Learning}
To handle missing modalities, a prevalent strategy is to leverage knowledge transfer or mutual learning between modalities, ensuring that models trained with incomplete data can benefit from knowledge acquired under complete modalities. These methods can be categorized based on their core mechanism: feature hallucination, knowledge distillation, and collaborative or adversarial learning. Early work includes feature hallucination approaches like HallNet~\cite{kampffmeyer2018urban}, which directly generates features of a missing modality. More commonly, knowledge distillation is employed, where a teacher model trained on complete data guides a student model for missing-modality inference. In remote sensing, DisOptNet~\cite{kang2022disoptnet} distills semantic knowledge from optical to SAR images for all-weather building segmentation, while DH-ADNet~\cite{li2021dynamic} uses dynamic hierarchical attention distillation for land cover classification. Beyond a simple teacher-student framework, collaborative or adversarial learning methods foster interaction among modalities. ACN~\cite{wang2021acn} employs an adversarial collaborative network for brain tumor segmentation, and MMANet~\cite{wei2023mmanet} utilizes margin-aware distillation and modality-aware regularization for incomplete multimodal learning. Furthermore, DMNet~\cite{zheng2021deep} introduces the concept of meta-sensors and prototypes to preserve sensor-invariant characteristics, generating specific networks when modalities are missing. While effective, these methods often focus on transferring or synthesizing knowledge without explicitly disentangling the underlying shared and unique factors across modalities, which is the focus of another line of research.

\subsection{Shared-Specific Feature Learning}
Another paradigm addresses modality absence by explicitly learning to  modality-shared and modality-specific representations. The underlying premise is that a robust, modality-invariant feature space can sustain performance when specific modalities are unavailable. These methods typically involve dedicated network branches and tailored loss functions to achieve separation. For instance, ShaSpec~\cite{wang2023multi} learns shared and specific features through auxiliary tasks like distribution alignment and domain classification, and synthesizes missing features by combining shared features from available modalities. A more advanced approach is presented by MetaRS~\cite{zhou2026remote}, which introduces a meta-modality perception module to extract invariant features and employs supervision on the multimodal feature covariance matrix to enforce disentanglement. It emphasizes operating in a meta-modality space to avoid corrupting modality-specific distributions during knowledge transfer. While this paradigm enhances theoretical robustness, a key challenge persists: the alignment process for learning shared representations is often applied uniformly across all pixels, making it vulnerable to severe class imbalance in complex scenes (e.g., remote sensing). This can lead to poor alignment and recognition for minority classes, a gap that our proposed Pixel-level Semantic sampling Alignment (PSA) strategy explicitly targets.

\subsection{Contrastive Learning}
The core idea of contrastive learning is to learn discriminative feature representations by pulling positive sample pairs closer and pushing negative sample pairs apart. Early representative works, such as SimCLR~\cite{chen2020simclr}, relied heavily on a large number of negative samples and very large training batch sizes to prevent feature collapse. To alleviate the reliance on large batches, MoCo~\cite{he2020momentum} introduced a momentum encoder and a dynamic queue to maintain a large number of negative samples. Subsequently, some studies began to challenge the view that negative samples are essential. For instance, SimSiam~\cite{chen2021exploring} revealed an alternating optimization mechanism that resembles Expectation-Maximization (EM) with a stop-gradient operation. In this mechanism, the stop-gradient operation forces the model to switch between two sub-problems, thereby avoiding the trivial solution of a constant output.

\section{Methodology}
\label{sec:method}
\subsection{Overall Architecture}

\begin{figure}[ht]
    \centering
    \includegraphics[width=1.0\textwidth]{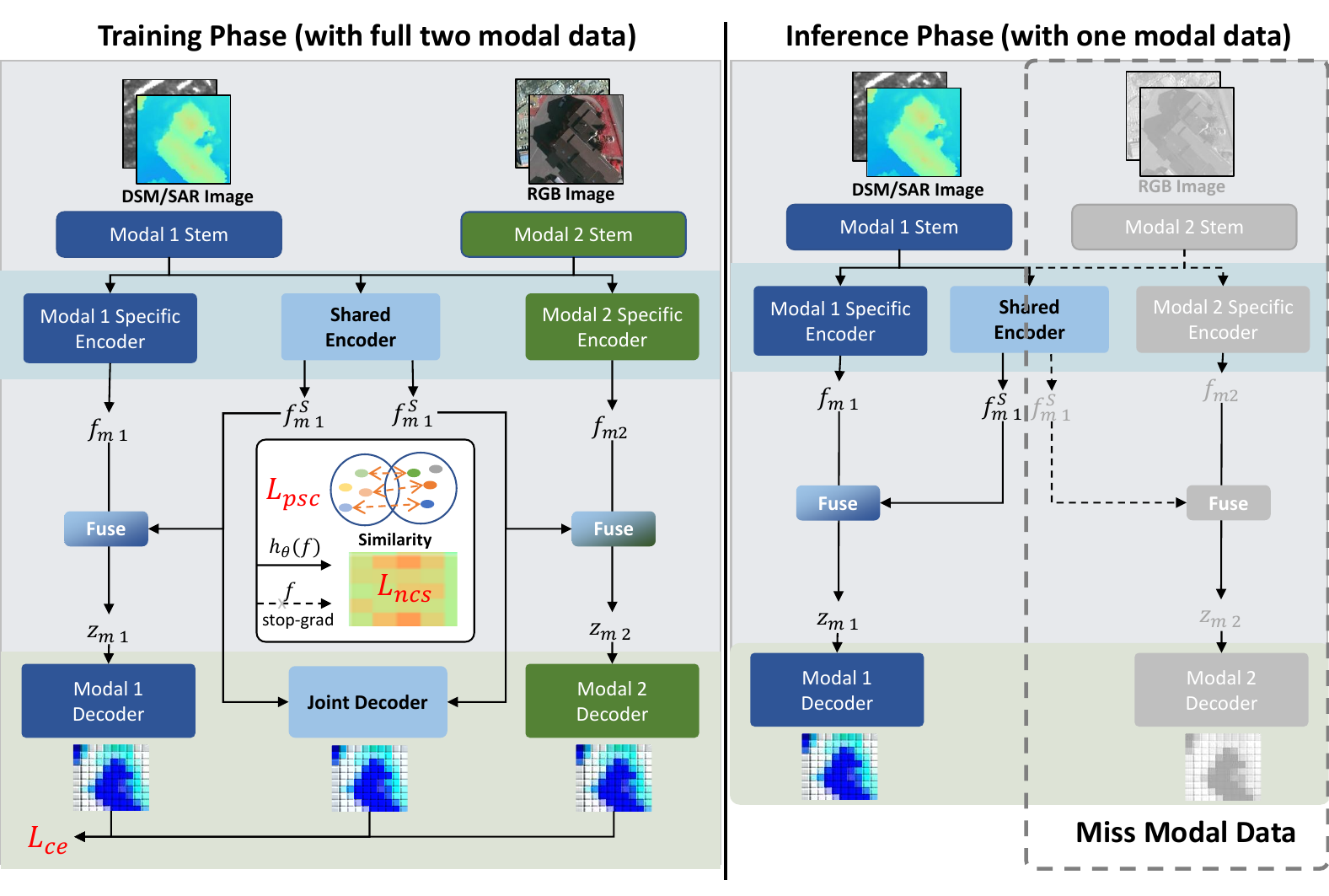} 
        \caption{Overview of the STARS (\textbf{S}hared-specific \textbf{T}ranslation and \textbf{A}lignment for missing-modality \textbf{R}emote \textbf{S}ensing Semantic Segmentation) framework.  During training, the architecture employs shared and modality-specific encoders to extract complementary features from multimodal data through cross-modality alignment and fusion. During inference, the learned universal feature representations enable high-precision performance under single-modality constraints.}
    \label{fig:framework}
\end{figure}

To address the issues of semantic alignment and feature extraction in remote sensing images under missing-modality scenarios, we propose the STARS (\textbf{S}hared-specific \textbf{T}ranslation and \textbf{A}lignment for missing-modality \textbf{R}emote \textbf{S}ensing Semantic Segmentation) framework, whose overall structure is shown in Figure \ref{fig:framework}. For heterogeneous modalities $x_{1}, x_{2}$ (e.g., SAR and optical, DSM and optical), we designs a shared-specific feature modeling architecture. The model first utilizes Modality Stems to perform initial downsampling and feature mapping on the input images, adapting them to the input channel requirements of the subsequent encoders. In the core encoding layers, STARS includes a Shared Encoder to capture inter-modality common semantic information and two Specific Encoders to preserve the physical imaging characteristics of their respective modalities, thereby learning modality-specific features.

In traditional multimodal feature alignment methods, to prevent the model from falling into a Trivial Solution (e.g., all outputs being predicted as zero vectors), it is typically necessary to carefully tune the weight of the alignment loss (e.g., some studies set it to extremely small values like 0.02~\cite{wang2023multi}). This high sensitivity to hyperparameters limits the model's robustness. Therefore, we introduce the bidirectional Translation Modules ($h_\theta$) and stop-gradient mechanism inspired by SimSiam~\cite{chen2021exploring}, which introduce asymmetry in the updates to break the symmetric balance between the two branches, effectively preventing the model from converging to a feature collapse solution of constant vectors.

To further enhance semantic consistency and overcome the issue of uneven distribution of remote sensing objects, STARS adopts a pixel-level sampling comparison alignment strategy during the training phase. By computing cross-modality consistency loss on sampled pixels, this strategy balances the distribution of positive and negative samples in the comparison process, thereby enforcing high-level semantic alignment of multimodal features.

\subsubsection{Fusion Module}

\begin{figure}[htb]
    \centering
    \includegraphics[width=0.4\linewidth]{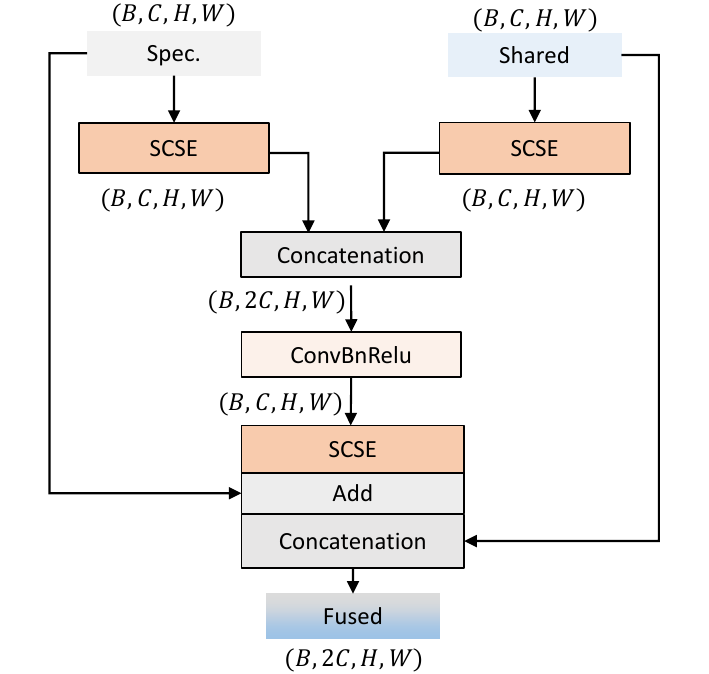}
    \caption{Detailed architecture of the Fusion Module. It integrates modality-specific and shared features using SCSE blocks, residual connections, and concatenation to preserve both distinct physical characteristics and common semantic information.}
    \label{fig:stars_fusion}
\end{figure}

The Fusion Module (Figure \ref{fig:stars_fusion}) is designed to effectively aggregate the learned modality-shared semantics ($F_{shared}$) and modality-specific details ($F_{spec}$), generating embeddings that are both semantically rich and modality-discriminative. At its core, the module employs a residual design and the SCSE (Spatial and Channel Squeeze and Excitation) attention mechanism \cite{roy2018concurrent} to enhance feature representation and preserve critical information.

As shown in Figure \ref{fig:stars_fusion}, the fusion process unfolds in several steps. First, both the specific and shared feature streams pass through independent SCSE blocks, which concurrently recalibrate their respective representations by emphasizing informative channels and spatial locations. The refined features are then concatenated along the channel dimension and processed by a Conv-BatchNorm-ReLU block to fuse cross-modality information and reduce dimensionality back to $C$. Crucially, to prevent the degradation of unique modality characteristics during fusion, we introduce a residual connection. The original specific features ($F_{spec}$) are added to the fused features after a second SCSE refinement. Finally, the original shared features ($F_{shared}$) are concatenated with this residual-enhanced output, yielding a final representation of shape $(B, 2C, H, W)$. This architecture ensures the decoder receives both the deeply integrated specific information and the preserved common semantic structure.

\begin{figure}[htb]
    \centering
    \includegraphics[width=0.5\linewidth]{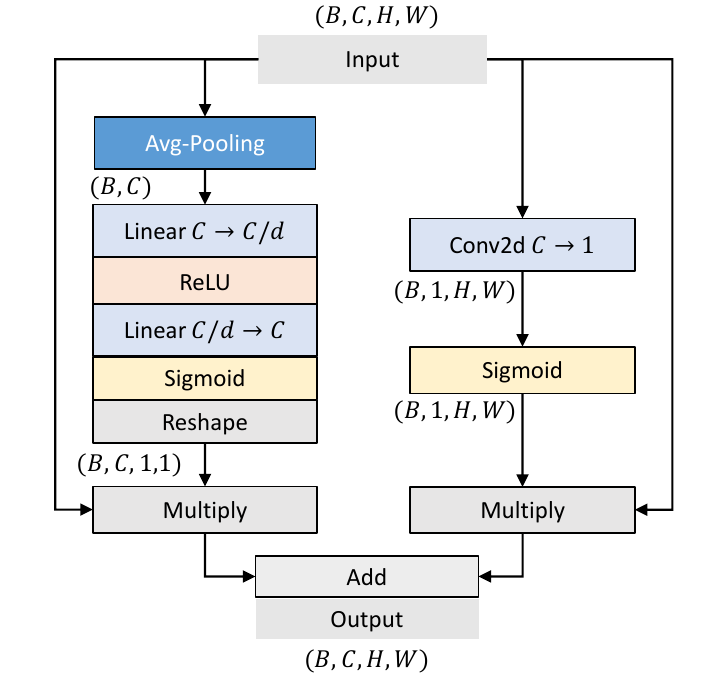}
    \caption{Structure of the Spatial and Channel Squeeze and Excitation (SCSE) block. It concurrently recalibrates features along the channel and spatial dimensions to enhance representational power.}
    \label{fig:scse}
\end{figure}

The SCSE block (Figure \ref{fig:scse}) serves as the core attention unit within the fusion module. It operates on an input feature map $U \in \mathbb{R}^{B \times C \times H \times W}$ through two parallel branches:
\begin{itemize}
    \item \textbf{Channel Squeeze and Excitation (cSE):} This branch applies global average pooling to the input, producing a channel descriptor vector of size $(B, C)$. It then passes through a bottleneck MLP (with layers: Linear $C \to C/d$, ReLU, Linear $C/d \to C$) followed by a Sigmoid activation, generating channel-wise attention weights. These weights are reshaped and multiplied with the input to emphasize informative channels.
    \item \textbf{Spatial Squeeze and Excitation (sSE):} The spatial branch compresses the channel dimension to 1 via a $1 \times 1$ convolution, followed by a Sigmoid function to produce a spatial attention map $\in \mathbb{R}^{B \times 1 \times H \times W}$. This map highlights relevant spatial locations .
\end{itemize}
The outputs of the cSE and sSE branches are fused via element-wise addition to produce the final recalibrated feature map.

\subsubsection{FPN Decoder}

In the decoder part, STARS is configured with three independent FPN (Feature Pyramid Network)~\cite{lin2017feature} decoders that perform supervision on the two single-modality branches and the fusion feature branch, respectively. This design ensures that the learned shared features retain modality-specific physical discriminability while also possessing rich fused semantic information.

\subsection{Stop-Gradient Based Asymmetric Feature Alignment}

In multimodal feature alignment, traditional Siamese architectures typically employ symmetric pathways to minimize feature distances. However, this symmetry often leads the optimization process toward a trivial solution (i.e., feature collapse), where the encoder outputs identical constant vectors for all inputs, causing a complete loss of semantic discriminability. While architectures like SimSiam have addressed this in self-supervised learning via asymmetric designs, applying them directly to remote sensing faces the challenge of significant distribution shifts between heterogeneous modalities (e.g., SAR vs. Optical). 

To address this, we propose an asymmetric alignment mechanism adapted for cross-modality scenarios. Instead of simple projection, we introduce a Bidirectional Translation Module to explicitly model the style transformation between modalities. By combining this with a stop-gradient operation, we break the symmetric equilibrium of the gradient updates. This design forces the network to learn meaningful semantic mappings rather than collapsing into a constant state, significantly reducing the model's sensitivity to hyperparameter tuning.

\subsubsection{Translation module}

To effectively align heterogeneous features within this asymmetric architecture, a simple linear projection is often insufficient to bridge the significant distributional gap between modalities. Therefore, we design a lightweight Translation Module, denoted as $h_\theta$. Unlike the standard predictor in SimSiam which mainly focuses on dimension matching, our Translation Module serves as a non-linear mapping function intended to translate the representation of one modality to approximate the latent feature distribution of the other. By incorporating this learnable module into the alignment branch, we introduce the structural asymmetry required to prevent collapse, and explicitly model the cross-modality domain shift, facilitating more robust semantic alignment.

Based on the implementation, the Translation Module $h_\theta$ transforms an input feature map $X \in \mathbb{R}^{C \times H \times W}$ through a sequential structure:

\begin{equation}
    h_\theta(X) = \text{Conv}_{1\times1}(\sigma(\text{IN}(\text{Conv}_{1\times1}(X))))
\end{equation}

where:
\begin{itemize}
    \item $\textbf{Conv}_{1\times1}$: A $1\times1$ convolutional layer with padding is first applied to aggregate local spatial context and encode neighborhood information.
    \item \textbf{IN (Instance Normalization)}: Unlike Batch Normalization which uses global batch statistics, we employ Instance Normalization. Since remote sensing modalities (e.g., SAR and Optical) possess vastly different statistical distributions and physical properties (styles), Instance Normalization helps to standardize the features per sample, removing modality-specific contrast information while preserving semantic content.
    \item $\sigma$ \textbf{(SiLU)}: The Sigmoid Linear Unit activation function ($x \cdot \text{sigmoid}(x)$)~\cite{elfwing2018sigmoid} is utilized to introduce non-linearity. SiLU provides a smooth, non-monotonic gradient flow compared to ReLU, which helps the network learn more complex patterns and stabilize the training process during the cross-modal translation.
    \item $\textbf{Conv}_{1\times1}$: Finally, a $1\times1$ convolution acts as a channel-wise projection layer to map the refined features to the target embedding space without altering the spatial resolution.
\end{itemize}

\subsubsection{Mathematical Definitions and Loss Function}

Under the STARS framework, given bimodal input data $x_1$ (e.g., SAR, DSM) and $x_2$ (RGB), a shared encoder extracts modality-specific features $f_{m1}$ and $f_{m2}$, respectively. To enable cross-modality mapping, a translation module $h^{m1 \Rightarrow m2}_{\theta}$ is defined to map the features of modality 1 into the feature space of modality 2, yielding the predicted representation $p_1$:

\begin{equation}
p_1 = h^{m1 \Rightarrow m2}_{\theta}(f_{m1})
\end{equation}

A translation module $h^{m2 \Rightarrow m1}_{\theta}$ is defined to map the features of modality 2 into the feature space of modality 1, yielding the predicted representation $p_2$:
\begin{equation}
p_2 = h^{m2 \Rightarrow m1}_{\theta}(f_{m2})
\end{equation}

When computing the alignment loss, the \textit{detach} operation is applied to block gradient backpropagation with respect to $f_{m1}$ and $f_{m2}$, treating them as fixed anchor points. The cross-modality translation loss $L_{ncs}$ is defined as:
\begin{equation}
L_{ncs} = - \frac{1}{2} \left[ \text{sim}(p_1, \text{stopgrad}(f_{m2})) + \text{sim}(p_2, \text{stopgrad}(f_{m1})) \right]
\end{equation}
where $\text{sim}$ denotes cosine similarity.

\subsubsection{Principle of the Asymmetric Mechanism}

The rationale for employing the stop-gradient operation is grounded in the theoretical analysis of SimSiam~\cite{chen2021exploring}, which hypothesizes that such Siamese architectures implicitly perform an Expectation-Maximization (EM) like alternating optimization. We extend this perspective to the cross-modality domain of STARS to explain how it avoids feature collapse while achieving semantic alignment.

Formally, let $\theta$ denote the parameters of the encoder and translation module, and let $\eta$ represent the latent semantic representation of the remote sensing scene. Following the analysis in~\cite{chen2021exploring}, the optimization objective can be formulated as minimizing a loss function $\mathcal{L}(\theta, \eta)$. The training process approximates an alternating minimization strategy:

\begin{enumerate}
    \item \textbf{E-step (Feature Approximation):} With the network parameters $\theta$ fixed, the model estimates the "true" semantic representation $\eta$. In the context of the stop-gradient branch, the frozen output (e.g., $\text{stopgrad}(f_{m2})$) serves as the current best estimate of $\eta$, acting as a stable semantic anchor.
    \item \textbf{M-step (Parameter Optimization):} With the target representation $\eta$ fixed, the network updates $\theta$ via SGD to minimize the distance between the predicted translation $p_1$ and $\eta$.
\end{enumerate}

\textbf{Analogy and Difference with SimSiam:} While STARS shares the underlying asymmetric optimization logic with SimSiam to prevent collapse, the nature of the alignment differs fundamentally due to the cross-modality scenario:

\textbf{Analogy (Structural Asymmetry):} Similar to SimSiam, STARS breaks the symmetry of the gradients. If both branches were updated simultaneously without stop-gradient, the system would rapidly minimize the loss by outputting a constant vector (trivial solution). The stop-gradient forces the predictor (Translation Module) to solve a harder task: predicting a moving target rather than negotiating a common collapse.
    
\textbf{Difference (Augmentation Invariance vs. Modality Invariance):} In SimSiam, the two views are generated via data augmentation (e.g., crop, color jitter) of the \textit{same} image. The "M-step" essentially learns \textit{augmentation invariance}. In contrast, STARS inputs heterogeneous data (e.g., SAR and Optical) with distinct physical distributions. Here, the "M-step" does not merely learn invariance but performs an explicit \textbf{domain translation}. The Translation Module $h_\theta$ is tasked with bridging the distribution gap (manifold matching) between modalities. Therefore, in STARS, the stop-gradient mechanism does not just stabilize training; it enforces the learning of \textit{semantic consistency} that is invariant to physical imaging mechanisms, making the alignment task significantly more challenging and substantive than single-modality contrastive learning.

\subsection{Pixel-level Semantic Alignment}

In remote sensing image segmentation tasks, land cover classes often exhibit significant class imbalance. Large-area classes (e.g., water bodies, forests, or background regions) dominate in terms of pixel count, while critical targets (e.g., buildings, roads) occupy only a small fraction. If cross-modality contrastive learning is applied directly to all pixels, the model tends to fit the feature distribution of majority classes, leading to misalignment or loss of discriminative information for minority classes.

\begin{figure}[ht]
    \centering
    \includegraphics[width=0.9\textwidth]{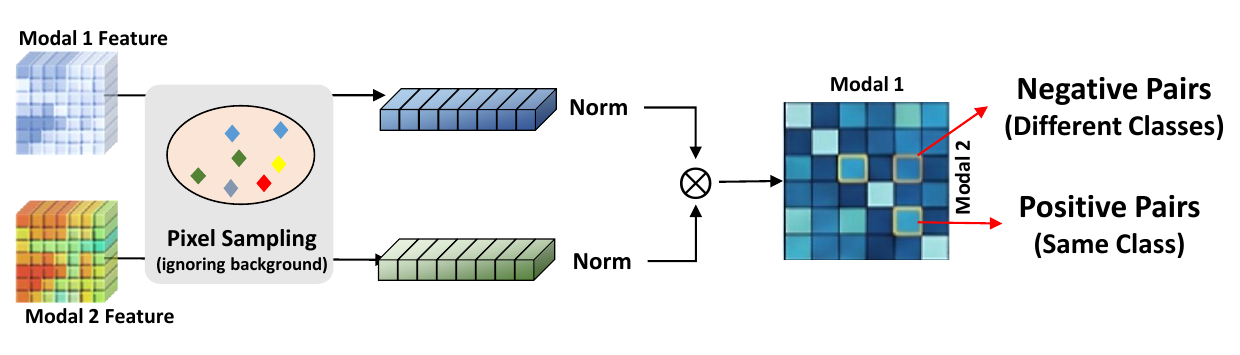} 
    \caption{Illustration of the Pixel-level Semantic Alignment (PSA) pipeline. This strategy employs an active sampling mechanism to extract class-balanced pixel samples from bimodal features.}

    \label{fig:sampleca}
\end{figure}

To address this issue, we propose a Pixel-level Semantic Alignment (PSA) strategy. This approach employs an active sampling mechanism to balance the distribution of positive and negative samples across different semantic classes, thereby enforcing precise alignment of multimodal features at the semantic level, as illustrated in Figure~\ref{fig:sampleca}.

\subsubsection{Semantic-balanced Sampling Mechanism}

In each training batch, the system first scans the ground truth to identify all valid semantic classes $C_{\text{valid}}$, excluding background or invalid regions. For each valid class $c \in C_{\text{valid}}$, it randomly samples a fixed number of pixel coordinate indices (e.g., $N = 32$ per class). If a given class has fewer than N pixels, sampling is repeated until N pixels are obtained; before sampling, the pixel coordinates for each class are randomly shuffled.

This mechanism equalizes the contribution of each class to the contrastive loss, regardless of its frequency in the full image. As a result, the model is encouraged to extract semantic information at multiple scales—such as building boundaries, road extents, and waterbody contours—with balanced attention, preventing minority-class targets from being overlooked during optimization.

\subsubsection{Cross-modality Contrastive Loss Based on Sampled Pixels}

After obtaining the sampled indices, feature vectors at the corresponding locations are extracted from the shared features $f_{m1}$ and $f_{m2}$, followed by $L_2$ normalization. The pixel-level semantic alignment loss $L_{psc}$ is computed as follows:

\begin{enumerate}
    \item \textbf{Similarity computation:} Compute the cosine similarity matrix between cross-modality sampled points. For the sampled pixels, the matrix element $\text{sim}(f_1^i, f_2^j)$ denotes the feature similarity between the $i$-th sampled point in modality 1 and the $j$-th sampled point in modality 2.
    
    \item \textbf{Positive pair identification:} Construct a semantic mask matrix. A cross-modality pair $(i, j)$ is considered positive if and only if the $i$-th pixel and the $j$-th pixel belong to the same semantic class. This criterion enforces not only alignment of co-located pixels across modalities but also intra-batch semantic consistency among pixels of the same land cover type.
    
    \item \textbf{Contrastive optimization:} Minimize a negative log-likelihood loss to pull together cross-modality pixel pairs with identical semantic labels and push apart those with different labels in the feature space.
\end{enumerate}

The pixel-level semantic alignment loss $L_{psc}$ is defined as:
\begin{equation}
L_{psc} = - \sum_{i \in I} \log \frac{\sum_{j \in P(i)} \exp(\text{sim}(f_1^i, f_2^j) / \tau)}{\sum_{k \in A(i)} \exp(\text{sim}(f_1^i, f_2^k) / \tau)}
\end{equation}

where $P(i)$ denotes the set of positive samples sharing the same semantic label as pixel $i$, $A(i)$ is the set of all sampled pixels, and $\tau$ is a temperature hyperparameter. The function $\text{sim}$ denotes cosine similarity.

\subsection{Joint Optimization and Multi-branch Supervision}

After completing the active cross-modality feature alignment, the framework performs end-to-end training using a multi-task joint loss function.

\subsubsection{Alignment Flow}

\begin{figure}[htp]
    \centering
    \includegraphics[width=1.0\linewidth]{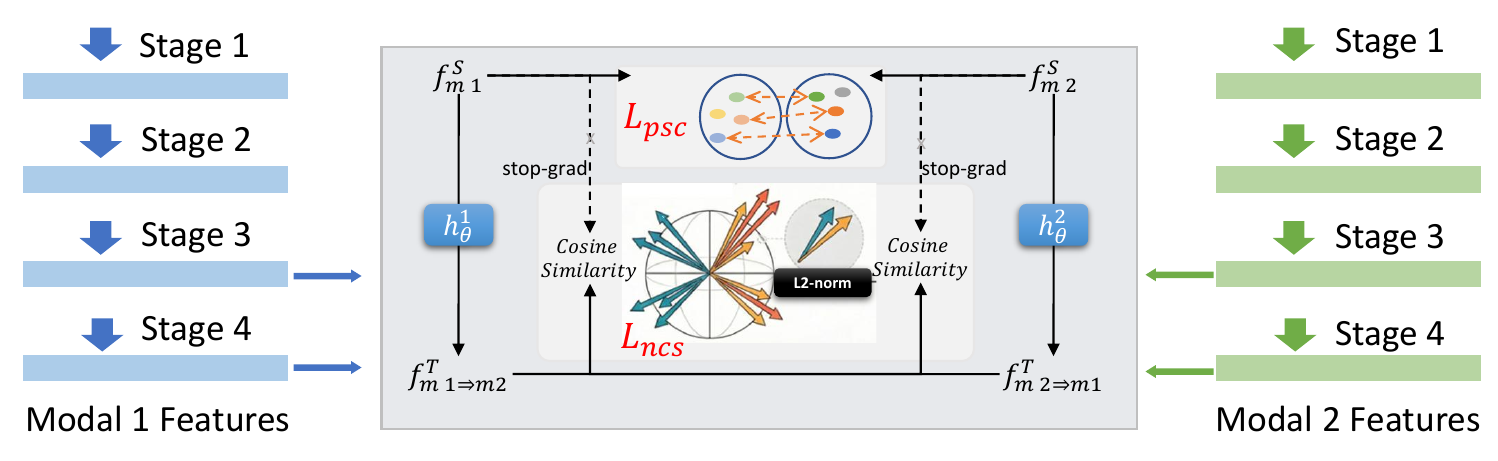}
    \caption{Deep features (Stage 3 \& Stage 4) undergo an alignment process jointly governed by the translation module $h_{\theta}$, the stop-gradient mechanism, and the combined constraints of the losses $L_{ncs}$ and $L_{psc}$.}

    \label{fig:alignment_flow}
\end{figure}

As shown in Figure~\ref{fig:alignment_flow}, the alignment module operates primarily on the high-level semantic features output by the encoder. Deep features (Stage 3 \& Stage 4) are aligned through a process jointly constrained by the translation module $h_{\theta}$, the stop-gradient mechanism, and the losses $L_{ncs}$ and $L_{psc}$.

\subsubsection{Multi-branch Decoder and Segmentation Loss}

The STARS framework employs three independent Feature Pyramid Network (FPN) decoders. As shown in Figure~\ref{fig:framework}, these decoders take as input the unimodal features and the fused shared features, respectively, and produce three segmentation predictions. The total supervised segmentation loss $L_{seg}$ is defined as the average of the cross-entropy losses from the three branches:
\begin{equation}
L_{seg} = \frac{1}{3} \left( L_{ce}(S_{m1}, Y) + L_{ce}(S_{m2}, Y) + L_{ce}(S_{fuse}, Y) \right)
\end{equation}
where $Y$ denotes the ground truth, and $S_{m1}$, $S_{m2}$, and $S_{fuse}$ represent the outputs of the three decoders, respectively.

\subsubsection{Joint Optimization Objective}

The overall optimization objective integrates the aforementioned translation consistency constraint and pixel-level semantic alignment constraint. By combining the task-specific segmentation loss with the modality alignment losses, the STARS framework simultaneously optimizes segmentation accuracy and enforces consistency in the cross-modality feature space. The final joint loss function $L_{\text{total}}$ is formulated as:
\begin{equation}
L_{\text{total}} = L_{seg} + \alpha L_{psc} + \beta L_{ncs}
\end{equation}
where $\alpha$ and $\beta$ are hyperparameters that control the relative contributions of the pixel-level semantic alignment loss $L_{psc}$ and the asymmetric alignment loss $L_{ncs}$, respectively.

\section{Experiments and Results}
\label{sec:experiments}

\subsection{Experimental Setup}
\label{subsec:setup}

\subsubsection{Implementation Details}

The proposed STARS is implemented using PyTorch, trained and tested on an NVIDIA V100 GPU with CUDA 11.7. For all datasets, the model is trained for a total of 80,000 training steps, with the first 1,000 steps dedicated to a linear warm-up phase. During this warm-up period, the learning rate increases linearly from $ 1 \times 10^{-6} $ to the initial value of $ 1 \times 10^{-4} $. Afterward, a cosine decay schedule is applied to gradually reduce the learning rate over the remaining steps. The batch size is set to 8, and the Adam optimizer is used with a weight decay of $ 1 \times 10^{-4} $. To ensure numerical stability during training, gradient clipping is applied with a maximum norm of 5.0. The shared encoder and two specific encoders adopted the ResNet-50~\cite{he2016deep} architecture.

\subsubsection{Dataset Description}
\label{sec:datasets}

\textbf{EarthMiss}
EarthMiss~\cite{zhou2026remote} is a high-resolution multimodal remote sensing land cover classification benchmark dataset, designed to simulate real-world scenarios with missing modalities (e.g., optical imagery obscured by clouds or fog). The dataset contains 3,355 pairs of 0.6-meter-resolution optical and synthetic aperture radar (SAR) images, covering 13 cities across five continents. EarthMiss includes eight common land cover classes, making it the most class-diverse high-resolution multimodal land cover dataset currently available.

\textbf{WHU-OPT-SAR}
The WHU-OPT-SAR dataset~\cite{li2022mcanet} is a 5-meter resolution dataset covering 51,448.56 $km^2$  in Hubei Province, China. It contains 100 pairs of co-registered optical (GF-1, with RGB and NIR bands) and SAR (GF-3) images, along with pixel-level annotations. The area features diverse terrains and vegetation types, with elevations ranging from 50 m to 3,000 m under a subtropical monsoon climate. All images are cropped to $512 \times 512$. Optical input used Nir-RG three-band, and SAR used single-band input.

\textbf{ISPRS Potsdam Dataset}  
The ISPRS Potsdam dataset is a popular benchmark for urban scene analysis, consisting of 38 true orthophoto images and nDSM maps. It includes four spectral bands (RGB and NIR) and five land cover categories. The dataset is split into 24 training and 14 test patches, with the latter further divided into 14 test and 1 validation patch. All images are cropped to $512 \times 512$ patches. Optical input used Nir-RG three-band, and nDSM used single-band input.

\subsubsection{Evaluation Metrics}

The performance of the proposed STARS model is evaluated using two standard metrics:

\textbf{Mean Intersection over Union (mIoU):}
The mIoU is computed as the average of the Intersection over Union (IoU) across all classes:
\begin{equation}
\text{mIoU} = \frac{1}{C} \sum_{i=1}^{C} \frac{TP_i}{TP_i + FP_i + FN_i}
\end{equation}
where $C$ is the number of classes, $TP_i$, $FP_i$, and $FN_i$ denote true positives, false positives, and false negatives for class $i$, respectively. mIoU is a widely-used metric for semantic segmentation that balances precision and recall.

\textbf{Mean F1-Score (mF1):}
The mF1 is the average F1-score across all classes:
\begin{equation}
\text{mF1} = \frac{1}{C} \sum_{i=1}^{C} 2 \times \frac{\text{Precision}_i \times \text{Recall}_i}{\text{Precision}_i + \text{Recall}_i}
\end{equation}
where Precision and Recall for class $i$ are calculated from the confusion matrix. The mF1 score provides a harmonic mean of precision and recall, offering a comprehensive assessment of model performance.

\subsubsection{Baseline Methods}

To evaluate the effectiveness of the proposed framework under modality-missing conditions, we compare against a diverse set of state-of-the-art baseline methods. These approaches represent various technical strategies for multimodal learning with incomplete inputs, including hallucination-based compensation, mutual learning, adversarial training, knowledge distillation, and cross-modality representation learning.

\textbf{Hallnet}~\cite{kampffmeyer2018urban}: This method introduces a hallucination network architecture that embeds all available modalities during training, enabling the model to implicitly compensate for missing modalities at test time and achieve multimodal fusion performance even with single-modality input.

\textbf{DML}~\cite{zhang2018deep}: Deep Mutual Learning (DML) proposes a collaborative learning strategy where multiple student networks teach each other during training, jointly improving their performance without relying on a pre-trained teacher network as in traditional knowledge distillation.

\textbf{DMNet}~\cite{zheng2021deep}: This approach presents a registration-free deep multisensor learning framework. By learning a meta-sensory representation and applying a difference alignment operation (DiffAlignOp), it addresses challenges of catastrophic forgetting and difficulties in precise data registration in multisensor collaboration.

\textbf{DH-ADNet}~\cite{li2021dynamic}: A dynamic hierarchical attention distillation network that incorporates multimodal collaborative instance selection (MSIS) and curriculum learning. Its dynamic hierarchical attention distillation module (DH-ADM) adaptively distills privileged information from multi-level features to handle missing modalities at test time.

\textbf{ACN}~\cite{wang2021acn}: An adversarial co-training network that couples full-modality and partial-modality learning processes. It minimizes the domain gap and recovers latent representations of missing modalities through entropy-based and adversarial knowledge learning modules.

\textbf{DisOptNet}~\cite{kang2022disoptnet}: This method transfers semantic knowledge from optical images to SAR segmentation networks. By analyzing cross-modality feature discrepancies and employing parallel convolutional branches, it leverages optical guidance to improve building segmentation in SAR imagery under adverse weather conditions.

\textbf{MMANet}~\cite{wei2023mmanet}: This framework supports incomplete multimodal learning via mask-aware distillation (MAD) and modality-aware regularization (MAR). It uses a teacher network to transfer comprehensive information and guides the deployable student network to adaptively balance weak modality combinations.

\textbf{ShaSpec}~\cite{wang2023multi}: This approach models both shared and modality-specific features (Shared-Specific Feature Modelling) to handle missing modalities. It relies on auxiliary tasks of distribution alignment and domain classification, offering a simple yet flexible architecture applicable to both classification and segmentation.

\textbf{RobustSAM}~\cite{chen2024robustsam}: An enhanced variant of the Segment Anything Model (SAM) that improves robustness on low-quality images with minimal additional parameters and computation, while preserving SAM’s promptability and zero-shot generalization capabilities.

\textbf{MetaRS}~\cite{zhou2026remote}: This method proposes a meta-modality representation model tailored for remote sensing land cover mapping under modality-unavailable scenarios. It aims to learn universal cross-modality feature representations to improve performance when certain modalities are missing.

\subsection{Comparison with state of the arts}

\subsubsection{Results on EarthMiss Dataset}

\begin{table}[htbp]
    \centering
    \caption{Quantitative evaluation results on the EarthMiss dataset. Results marked with $^*$ are cited from~\cite{zhou2026remote}. \textbf{S} denotes the SAR modality and \textbf{O} denotes the optical modality. The best result in each column is highlighted with \colorbox{red!10}{\textbf{bold}}, and the second-best result is indicated by \colorbox{black!10}{\underline{underlined}}.}

    \label{tab:earthmiss}
    \renewcommand{\arraystretch}{1.4}
    \begin{adjustbox}{width=\textwidth} 
    \begin{tabular}{l|cc|cccccccc|cc}
        \toprule
        \textbf{Method} & \textbf{Train} & \textbf{Test} & \textbf{Backgr.} & \textbf{Build.} & \textbf{Road} & \textbf{Water} & \textbf{Barren} & \textbf{Forest} & \textbf{Agri.} & \textbf{Playg.} & \textbf{$mIoU$} & \textbf{$mF1$} \\ 
        \midrule
        Baseline (Concat) & S+O & S+O & 36.21 & 60.13 & 48.72 & 93.81 & 53.55 & 45.72 & 40.12 & 16.85 & 49.39 & 63.71 \\
        Baseline-OPT & O & O & 35.92 & 60.70 & 44.18 & 94.81 & 50.31 & 44.81 & 41.32 & 21.07 & 49.14 & 62.76 \\
        Baseline-SAR & S & S & 28.72 & 18.10 & 13.42 & 83.12 & 13.14 & 28.34 & 0.00 & 2.79 & 23.45 & 31.42 \\
        \midrule
        Fine-tuning & O & S &  \cellcolor{black!10} \underline{30.81} & 37.07 & 18.08 & 72.86 & 12.82 & 27.01 & 0.00 & 2.43 & 25.13 & 35.10 \\
        $^\star$Hallnet~\cite{kampffmeyer2018urban} & S+O & S & 28.87 & 28.85 & 15.08 & 83.90 & 10.79 & 27.61 & 0.00 & 0.45 & 24.45 & 33.85 \\
        $^\star$DisOptNet~\cite{kang2022disoptnet} & S+O & S & 30.01 & 21.22 & 13.33 & 81.69 & 14.54 & 33.86 & 0.00 & 1.61 & 24.55 & 33.85 \\
        $^\star$DH-ADNet~\cite{li2021dynamic} & S+O & S & \cellcolor{red!10} \textbf{37.60} & 31.70 & 13.35 & 71.26 & 27.88 & \cellcolor{black!10} \underline{35.52} & 0.00 & 0.00 & 25.91 & 37.78 \\
        $^\star$ACN~\cite{wang2021acn} & S+O & S & 31.50 & 9.70 & 3.70 & 77.66 & 27.15 & 23.92 & 0.00 & 0.00 & 21.66 & 30.13 \\
        $^\star$ShaSpec~\cite{wang2023multi} & S+O & S & 30.57 & 36.84 & 16.74 & 76.26 & 23.44 & 19.58 & 0.00 & 0.10 & 24.44 & 35.80 \\
        $^\star$DML~\cite{zhang2018deep} & S+O & S & 30.56 & 16.90 & 11.84 & 70.90 & 11.76 & 26.63 & 0.00 & 0.50 & 21.14 & 30.50 \\
        $^\star$MMANet~\cite{wei2023mmanet} & S+O & S & 28.57 & 38.98 & 19.02 & 69.29 & 25.06 & 27.87 & 0.00 & 0.40 & 26.15 & 37.67 \\
        $^\star$RobustSAM~\cite{chen2024robustsam} & S+O & S & 22.94 & 33.87 & 2.40 & 76.09 & 17.10 & 8.50 & 0.00 & 0.00 & 20.11 & 27.98 \\
        $^\star$DMNet~\cite{zheng2021deep} & S+O & S & 27.52 & 33.97 & 13.33 & 64.29 & 21.00 & 35.17 & 0.00 & 0.00 & 24.41 & 35.35 \\
        $^\star$MetaRS~\cite{zhou2026remote} & S+O & S & 26.23 & \cellcolor{black!10} \underline{47.00} & \cellcolor{black!10} \underline{23.67} & \cellcolor{black!10} \underline{85.62} & \cellcolor{red!10} \textbf{31.82} & 35.26 & \cellcolor{black!10} \underline{0.29} & \cellcolor{black!10} \underline{4.26} & \cellcolor{black!10} \underline{31.77} & \cellcolor{black!10} \underline{41.56} \\
        \midrule
        \textbf{STARS (Ours)} & S+O & S & 22.72 & \cellcolor{red!10} \textbf{48.12} & \cellcolor{red!10} \textbf{26.32} & \cellcolor{red!10} \textbf{85.79} & \cellcolor{black!10} \underline{30.72} & \cellcolor{red!10} \textbf{37.71} & \cellcolor{red!10} \textbf{1.71} & \cellcolor{red!10} \textbf{4.27} & \cellcolor{red!10} \textbf{33.04} & \cellcolor{red!10} \textbf{46.22} \\
        \rowcolor{green!10} \textit{w.r.t.} baseline-SAR & & & +1.00 & +30.02 & +12.90 & +2.67 & +17.58 & +9.37 & +1.71 & +1.48 & +9.60 & +13.80 \\
        \bottomrule
    \end{tabular}
    \end{adjustbox}
\end{table}

\begin{figure}[htp]
    \centering
    \includegraphics[width=1.0\linewidth]{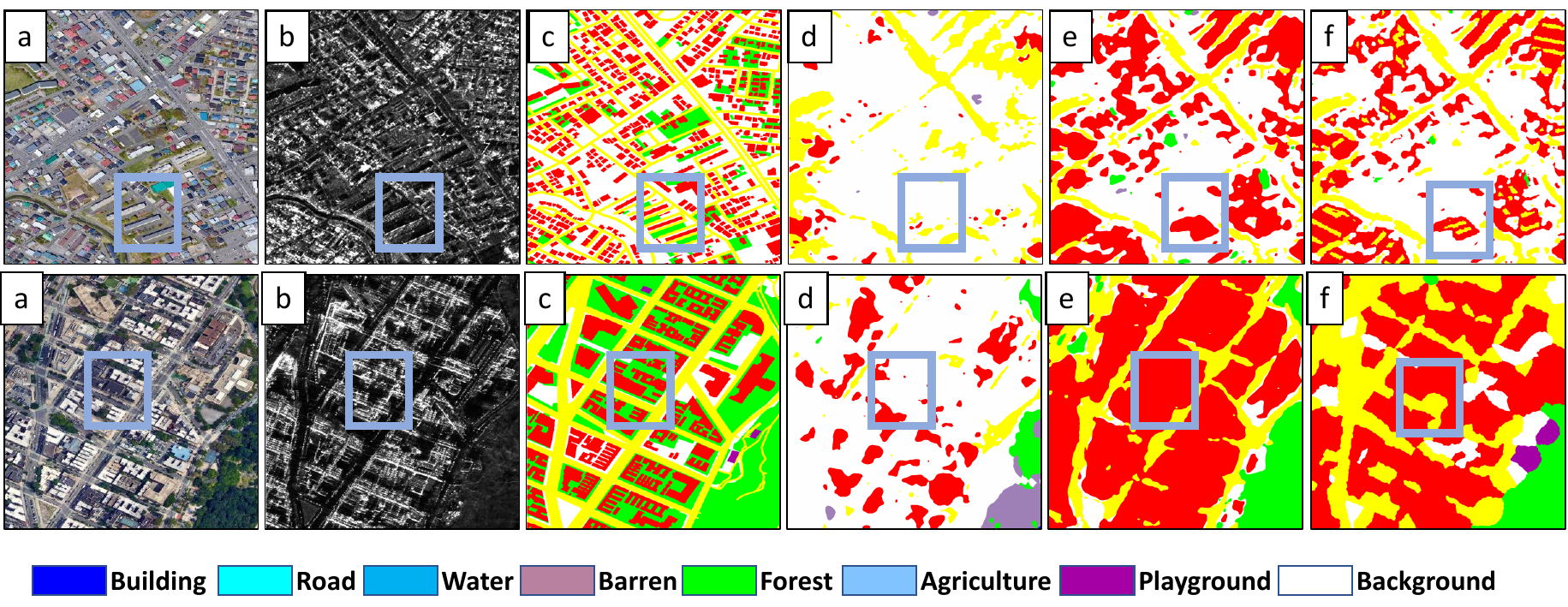}
    \caption{Qualitative segmentation results on the EarthMiss dataset. Two representative scenes are shown: (a) original optical image (reference), (b) corresponding SAR image (test input), (c) ground truth, (d) prediction from Baseline-SAR, (e) prediction from MetaRS, and (f) prediction from STARS (Ours). The legend at the bottom indicates different land cover classes. Regions within blue boxes highlight the models’ ability to capture building boundaries and road details.}

    \label{fig:missearth-demo}
\end{figure}

As shown in Table~\ref{tab:earthmiss}, the proposed STARS framework achieves significantly better quantitative results than existing methods on the EarthMiss dataset. Under the challenging test scenario using only single-modality SAR images as input, STARS attains the highest $mIoU$ of 33.04\% and $mF1$ of 46.22\%, representing improvements of 9.60\% and 13.80\% over the Baseline-SAR model, respectively.

Qualitative comparisons in Figure~\ref{fig:missearth-demo} further demonstrate STARS’ superior capability in recovering fine land cover details. Notably, STARS achieves the highest IoU scores of 48.12\% for Building and 26.32\% for Road. However, it is worth noting that our method exhibits a performance drop in the Background class (22.72\%) compared to baselines like DH-ADNet (37.60\%). This phenomenon offers a critical insight into the trade-off mechanism of our proposed Pixel-level Semantic Alignment (PSA). In standard training, the Background class typically dominates the pixel count, biasing the model towards it. By enforcing a class-balanced sampling strategy, STARS deliberately down-weights the contribution of the background to focus on optimizing minority but semantically significant classes (e.g., Agriculture and Roads). While this sacrifices some precision on the heterogeneous background, it yields substantial gains in identifying specific foreground targets, which is often more valuable in practical remote sensing applications.

As highlighted in the blue boxes of Figure~\ref{fig:missearth-demo}, the Baseline-SAR prediction (d) suffers from severe fragmentation and missed detections due to speckle noise in SAR imagery. Although MetaRS (e), the second-best method, captures approximate object outlines, it still lacks precision at boundaries. In contrast, the segmentation result from STARS (f) closely matches the ground truth (c), accurately recovering the regular geometric shapes of buildings and preserving the topological connectivity of roads.

\subsubsection{Results on WHU-OPT-SAR Dataset}

\begin{table*}[htbp]
    \centering
    \caption{Quantitative evaluation results on the WHU-OPT-SAR dataset. \textbf{S} denotes the SAR modality and \textbf{O} denotes the optical modality. The best result in each column is highlighted with \colorbox{red!10}{\textbf{bold}}, and the second-best result is indicated by \colorbox{black!10}{\underline{underlined}}.}
    \label{tab:results_whuoptsar}
    \renewcommand{\arraystretch}{1.4} 
    \resizebox{\textwidth}{!}{ 
    \begin{tabular}{l|cc|ccccccc|cc}
        \toprule
        Method & Train & Test & Farmland & City & Village & Water & Forest & Road & Others & $mIoU$ & $mF1$ \\
        \midrule
        Baseline (Concat) & S+O & S+O & 67.55 & 57.51 & 48.83 & 61.50 & 73.84 & 37.00 & 17.01 & 51.89 & 66.17 \\
        Baseline-OPT & O & O & 65.76 & 55.93 & 49.23 & 59.10 & 76.39 & 43.44 & 9.44 & 51.33 & 65.84 \\
        Baseline-SAR & S & S & 33.68 & 26.49 & 10.06 & 53.21 & 45.87 & 12.64 & 14.86 & 28.12 & 41.32 \\
        \midrule
        Fine-tuning & O & S & 33.11 & 31.83 & 21.85 & 41.18 & 58.74 & 12.51 & 14.06 & 30.47 & 44.34 \\
        DisOptNet~\cite{kang2022disoptnet} & S+O & S & \cellcolor{red!10} \textbf{34.81} & 31.27 & 19.50 & 41.81 & 57.63 & 11.82 & \cellcolor{red!10} \textbf{15.55} & 30.34 & 44.03 \\
        DH-ADNet~\cite{li2021dynamic} & S+O & S & \cellcolor{black!10} \underline{34.81} & 30.62 & 20.60 & 41.32 & 56.23 & 12.67 & \cellcolor{black!10} \underline{15.50} & 30.26 & 44.35 \\
        ShaSpec~\cite{wang2023multi} & S+O & S & 33.43 & 35.00 & \cellcolor{black!10} \underline{24.00} & 45.03 & 46.36 & 15.42 & 7.83 & 29.58 & 41.61 \\
        MMANet~\cite{wei2023mmanet} & S+O & S & 34.12 & 34.17 & \cellcolor{red!10} \textbf{24.36} & 45.15 & 44.76 & 15.63 & 9.88 & 29.72 & 41.67 \\
        RobustSAM~\cite{chen2024robustsam} & S+O & S & 34.16 & 32.52 & 15.41 & 48.50 & 13.79 & 1.95 & 12.19 & 22.64 & 34.61 \\
        DMNet~\cite{zheng2021deep} & S+O & S & 32.09 & 19.69 & 13.45 & 50.21 & 53.48 & 11.92 & 10.10 & 27.28 & 39.40 \\
        MetaRS~\cite{zhou2026remote} & S+O & S & 33.75 & \cellcolor{black!10} \underline{34.74} & 20.93 & \cellcolor{black!10} \underline{51.83} & \cellcolor{black!10} \underline{59.09} & \cellcolor{black!10} \underline{16.85} & 8.61 & \cellcolor{black!10} \underline{32.26} & \cellcolor{black!10} \underline{45.04} \\
        \midrule
        STARS (Ours) & S+O & S & 33.71 & \cellcolor{red!10} \textbf{39.97} & 21.37 & \cellcolor{red!10} \textbf{54.39} & \cellcolor{red!10} \textbf{62.71} & \cellcolor{red!10} \textbf{17.41} & 8.91 & \cellcolor{red!10} \textbf{34.07} & \cellcolor{red!10} \textbf{46.68} \\
        \rowcolor{green!15} \textit{w.r.t.} baseline-SAR & - & - & +0.03 & +13.48 & +11.31 & +1.18 & +16.84 & +4.77 & -5.95 & +5.95 & +5.36 \\
        \bottomrule
    \end{tabular}
    }
\end{table*}

\begin{figure}[htp]
    \centering
    \includegraphics[width=1.0\linewidth]{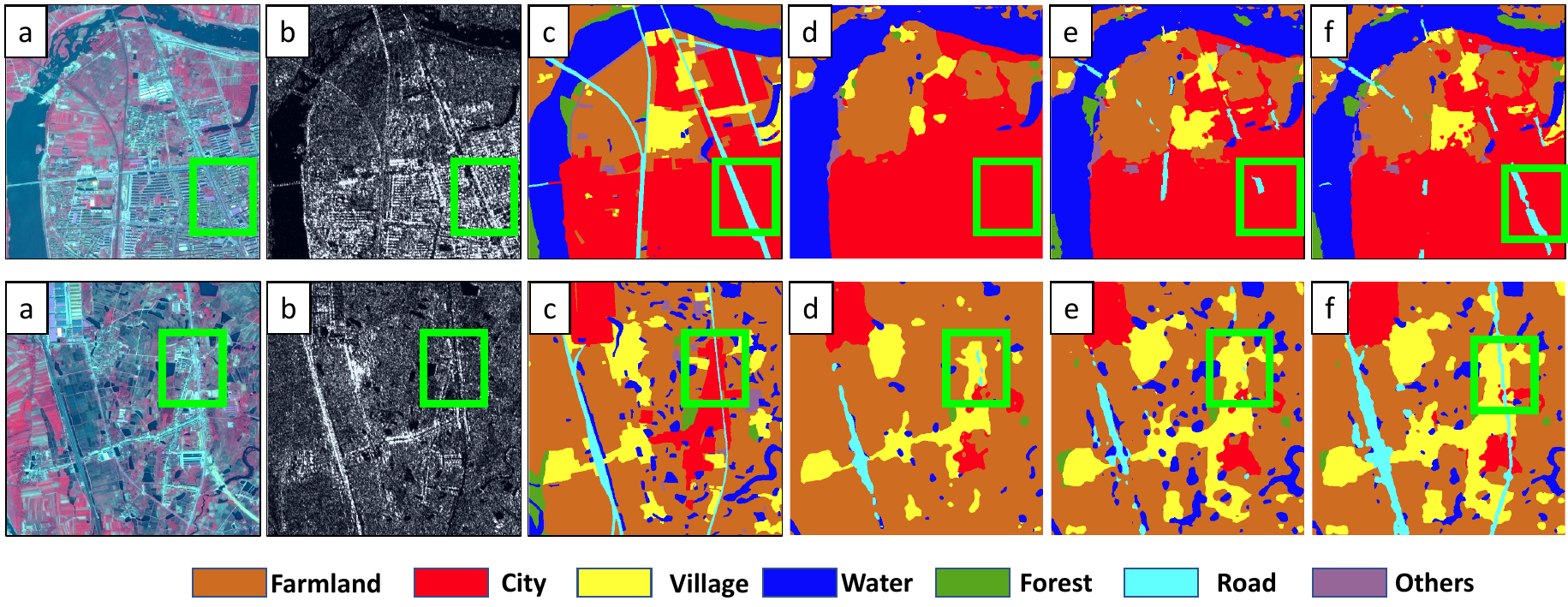}
    \caption{Qualitative comparison on the WHU-OPT-SAR dataset under the test setting with the SAR modality only as input. (a) Reference optical image; (b) SAR image for testing; (c) Ground truth (GT); (d) Prediction from Baseline-SAR; (e) Prediction from MetaRS; (f) Prediction from STARS (Ours). The legend at the bottom indicates different land cover classes. The green boxes highlight the performance of our method in capturing linear structures such as roads.}
    \label{fig:whuoptsar-demo}
\end{figure}

As shown in Table~\ref{tab:results_whuoptsar}, quantitative results on the WHU-OPT-SAR dataset demonstrate that STARS achieves state-of-the-art performance across most evaluation metrics, with an $mIoU$ of 34.07\% and an $mF1$ score of 46.68\%, significantly outperforming the current second-best method, MetaRS. Notably, on complex land cover classes such as Urban, Village, and Forest, STARS improves IoU by 13.48\%, 11.31\%, and 16.84\%, respectively, compared to the Baseline-SAR model. This substantial gain is primarily attributed to the pixel-level semantic sampling alignment (PSA) strategy in STARS.  We also observe a similar trend in the Others category, where STARS (8.91\%) scores lower than DisOptNet (15.55\%). The Others class usually contains ambiguous textures lacking distinct semantic patterns. Since our asymmetric alignment module relies on learning strong, distinct semantic representations from the Optical modality to guide the SAR branch, it is less effective on categories with weak semantic definitions. Conversely, for classes with strong structural priors like Roads and Buildings, the alignment mechanism successfully transfers the topological continuity from Optical to SAR features, resulting in the observed sharp boundaries shown in Figure~\ref{fig:whuoptsar-demo}.

This advantage is visually confirmed in Figure~\ref{fig:whuoptsar-demo}. Within the green boxed regions, the inherent speckle noise and geometric distortions in SAR images cause both Baseline-SAR (d) and MetaRS (e) to fail in preserving the continuous physical structure of narrow roads, resulting in fragmented or misclassified predictions. In contrast, STARS (f) produces more complete and spatially coherent road segments (shown in cyan) with sharper boundaries, consistent with its leading $IoU$ performance on the Road class reported in Table~\ref{tab:results_whuoptsar}. Taken together, both quantitative metrics and visual fidelity confirm that STARS effectively transfers rich semantic priors from the optical modality into the SAR feature space, demonstrating its capability in addressing highly challenging remote sensing segmentation tasks under severe modality degradation.

\subsubsection{Results on ISPRS Potsdam Dataset}

\begin{table}[htbp]
  \centering
  \caption{Quantitative evaluation results on the ISPRS Potsdam dataset. \textbf{S} denotes the DSM modality and \textbf{O} denotes the optical modality. The best result in each column is highlighted with \colorbox{red!10}{\textbf{bold}}, and the second-best result is indicated by \colorbox{black!10}{\underline{underlined}}.}

  \label{tab:results_potsdam}
    \renewcommand{\arraystretch}{1.4} 
    \resizebox{\textwidth}{!}{ 
  \begin{tabular}{l|cc|ccccc|cc}
    \toprule
    \textbf{Methods} & \textbf{Train} & \textbf{Test} & \textbf{Imperv.} & \textbf{Build.} & \textbf{Low Veg.} & \textbf{Tree} & \textbf{Car} & \textbf{$mIoU$} & \textbf{$mF1$} \\
    \midrule
    Baseline (Concat) & O+D & O+D & 85.29 & 89.74 & 75.31 & 76.64 & 73.48 & 80.09 & 88.79 \\
    baseline-DSM      & D & D & 82.71 & 86.16 & 61.27 & 71.34 & 42.73  & 68.84 & 71.76\\
    baseline-RGB      & O & O & 83.91 & 89.13 & 73.63 & 75.02 & 75.41& 79.42 & 87.27  \\
    \midrule
    Fine-tuning       & O+D & O & 84.31 & 91.51 & 73.32 & 75.05 & 74.03& 79.64 & 87.08  \\
    DisOptNet~\cite{kang2022disoptnet} & O+D & O & 84.22 & 88.12 & \cellcolor{black!10} \underline{75.34} & 76.21 & 74.23 & 79.72 & 87.16  \\
    MMANet~\cite{zheng2021deep}       & O+D & O & 84.76 & 90.69 & 72.36 & 74.79 & 74.07& 79.72 & 87.88  \\
    MetaRS~\cite{zhou2026remote}      & O+D & O & \cellcolor{red!10} \textbf{86.64} & \cellcolor{black!10} \underline{91.67} & 75.08 & \cellcolor{black!10} \underline{76.22} & \cellcolor{black!10} \underline{74.53} & \cellcolor{black!10} \underline{80.83} & \cellcolor{black!10} \underline{88.63} \\
    \midrule
    \textbf{STARS (Ours)} & O+D & O & \cellcolor{black!10} \underline{86.17} & \cellcolor{red!10} \textbf{91.69} & \cellcolor{red!10} \textbf{76.96} & \cellcolor{red!10} \textbf{78.57} & \cellcolor{red!10} \textbf{75.42}& \cellcolor{red!10} \textbf{81.76} & \cellcolor{red!10} \textbf{89.28}  \\
    \rowcolor{green!15} \textit{w.r.t.} baseline-RGB  & - & - & +2.26 & +2.54 & +3.33 & +3.55 & +0.01& +2.34  & +2.01 \\
    \bottomrule
  \end{tabular}
  }
\end{table}

\begin{figure}[htp]
    \centering
    \includegraphics[width=1.0\linewidth]{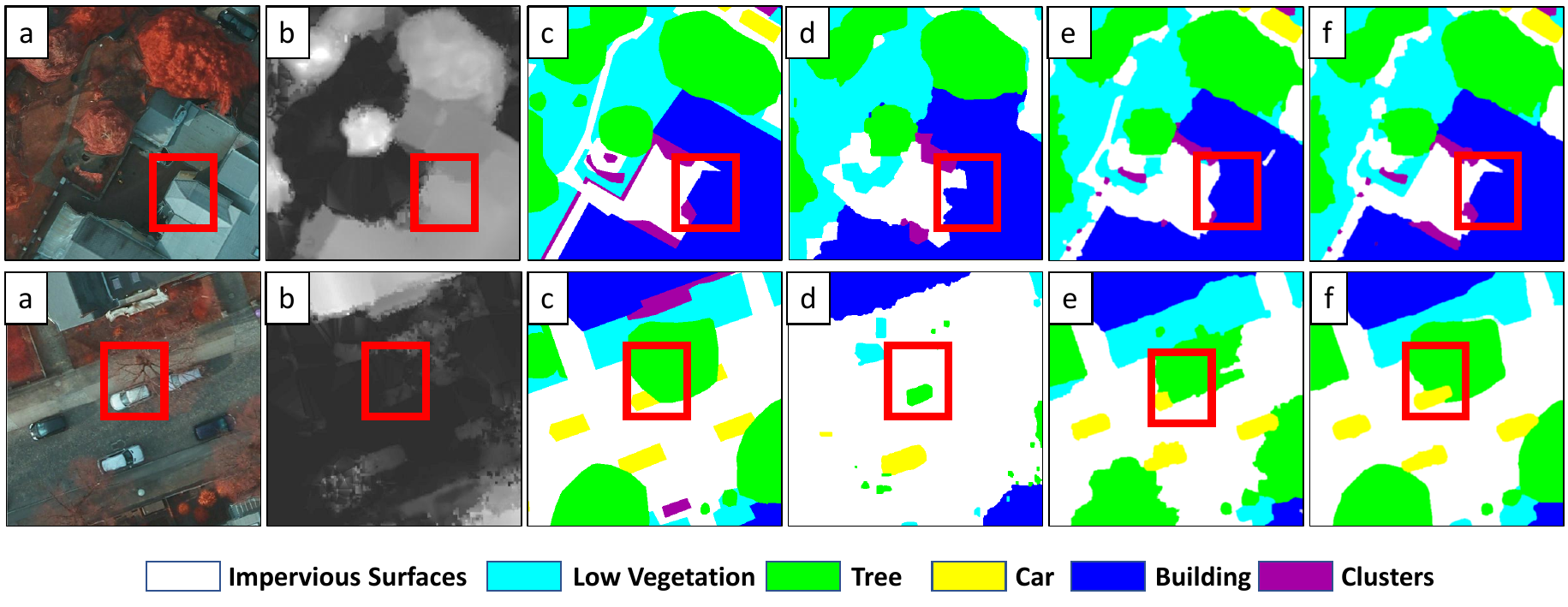}
    \caption{Qualitative comparison of prediction results on the ISPRS Potsdam dataset under the test setting with optical imagery as the only input. (a) Optical image (IRRG); (b) DSM image (for reference only); (c) Ground truth (GT); (d) Prediction from Baseline-RGB; (e) Prediction from MetaRS; (f) Prediction from STARS (Ours). The legend at the bottom indicates different land cover classes, and the red boxes highlight representative regions where the methods exhibit noticeable performance differences.}
    \label{fig:pot-demo}
\end{figure}

Table~\ref{tab:results_potsdam} presents the quantitative evaluation results of various methods on the ISPRS Potsdam dataset. Under the modality-missing setting—where models are trained with both optical (O) and DSM (D) modalities but tested using only the optical modality—STARS outperforms all competing methods across key metrics. Specifically, STARS achieves an $mIoU$ of 81.76\% and an $mF1$ of 89.28\%, representing a 2.34\% $mIoU$ improvement over Baseline-RGB, which is trained solely on optical imagery. In terms of per-class performance, the most notable gains occur in vegetation categories: IoU for Low Vegetation and Tree increases by 3.33\% and 3.55\%, respectively. Additionally, STARS attains the best results for Building (91.69\% IoU) and Car (75.42\% IoU).

Figure~\ref{fig:pot-demo} shows qualitative predictions on the test set. The visual improvements align consistently with the quantitative gains in Table~\ref{tab:results_potsdam}. In the red-boxed region of the first row in Figure~\ref{fig:pot-demo}, the absence of height information leads Baseline-RGB (d) and MetaRS (e) to misclassify boundaries between Buildings (blue) and Impervious Surfaces (white). In contrast, STARS (f) produces building outlines that closely match the ground truth (c). In the second row, STARS accurately detects shadow-occluded Trees (green) and small Cars (yellow), whereas Baseline-RGB exhibits clear missed detections. These observations indicate that, through the asymmetric feature alignment mechanism in STARS, the model effectively leverages geometric priors from the DSM modality. 

 Specifically, the significant improvements in "Building" (+2.54\%) and "Tree" (+3.55\%)—classes heavily dependent on height information—suggest that the Translation Module has successfully encoded the latent height semantics from the missing DSM modality into the Optical feature space. Unlike color-dependent classes (e.g., Impervious surfaces) where the gain is marginal, the model shows distinct advantages in distinguishing objects with similar spectral signatures but different elevations (e.g., trees vs. low vegetation), validating the efficacy of the cross-modality hallucination capability.

\subsection{Ablation studies}

In this section, we conduct a series of ablation studies to validate the effectiveness of each core component in the STARS model and to analyze its sensitivity to key hyperparameters and sampling strategies. All experiments are performed on the EarthMiss and WHU-OPT-SAR datasets, with mean Intersection over Union ($mIoU$) used as the evaluation metric.

\subsubsection{Effectiveness of Core Modules}

\begin{table}[htbp]
    \centering
    \caption{Analysis of the impact of different module components on model performance. Here, trans denotes the bidirectional translation module, $L_{ncs}$ refers to the asymmetric alignment loss based on stop-gradient, and $L_{psc}$ denotes the pixel-level semantic sampling alignment loss. The evaluation metric is $mIoU$.}

    \label{tab:ab_coremodules}
    \begin{tabular}{l|ccc|cc|cc}
        \toprule
        Method & trans & $L_{ncs}$ & $L_{psc}$ & EarthMiss &  & WHU-OPT-SAR &  \\
        \midrule
        Baseline-SAR & & & & 23.45 & -- & 28.12 & -- \\
        exp-1 & & \checkmark & & 30.71 & \textcolor{green!60!black}{+7.26} & 31.52 & \textcolor{green!60!black}{+3.40} \\
        exp-2 & \checkmark & \checkmark & & 31.17 & \textcolor{green!60!black}{+7.72} & 32.04 & \textcolor{green!60!black}{+3.92} \\
        exp-3 & & \checkmark & \checkmark & 32.79 & \textcolor{green!60!black}{+9.34} & 32.89 & \textcolor{green!60!black}{+4.77} \\
        STARS & \checkmark & \checkmark & \checkmark & 33.04 & \textcolor{green!60!black}{+9.60} & 34.07 & \textcolor{green!60!black}{+5.95} \\
        \bottomrule
    \end{tabular}
\end{table}

To evaluate the specific contribution of each component in the STARS framework to cross-modality feature alignment, we conduct ablation experiments as shown in Table~\ref{tab:ab_coremodules}. As described in the methodology, the core of STARS lies in addressing semantic deficiency in SAR imagery through asymmetric alignment loss ($L_{ncs}$) and pixel-level semantic sampling alignment loss ($L_{psc}$).

As observed in Table~\ref{tab:ab_coremodules}, the baseline model (Baseline-SAR), which uses only a standard encoder without cross-modality guidance, achieves relatively low performance due to the absence of semantic priors from the optical modality. Introducing $L_{ncs}$ (exp-1) yields significant improvements of 7.26\% on EarthMiss and 3.40\% on WHU-OPT-SAR, confirming the effectiveness of the asymmetric alignment strategy—specifically, that using optical features as fixed anchors to guide SAR feature learning is beneficial.

Building upon this, the addition of the bidirectional translation module (trans) and the pixel-level semantic sampling alignment loss ($L_{psc}$) (exp-2 and exp-3) further enhances the model’s ability to capture complex land cover semantics. Ultimately, the full STARS model achieves the highest $mIoU$ on both datasets (33.04\% on EarthMiss and 34.07\% on WHU-OPT-SAR), demonstrating the synergistic effect of translation consistency and semantic-level contrastive alignment in cross-modality representation learning.

\subsubsection{Impact of Loss Weights and Gradient Detachment}

\begin{table}[htbp]
    \centering
    \caption{Sensitivity analysis of the asymmetric alignment loss weight $\beta$ and effectiveness comparison of the gradient detachment (detach) operation. Experiments are conducted on the EarthMiss dataset with $mIoU$ as the evaluation metric.}

    \label{tab:beta}
    \begin{tabular}{ccc}
        \toprule
        $\beta$ & w/ detach & w/o detach \\
        \midrule
        0.02 & \cellcolor{red!10}{32.71} & \cellcolor{red!10}{32.46} \\
        0.1  & \cellcolor{red!10}{32.65} & 31.11 \\
        0.2  & \cellcolor{red!20}{33.04} & 30.81 \\
        0.5  & \cellcolor{red!10}{32.02} & 30.90 \\
        0.8  & 31.18 & 30.77 \\
        \bottomrule
    \end{tabular}
\end{table}

We further investigate the impact of loss weights $\beta$ and $\alpha$ on model performance. In particular, we validate the necessity of the gradient detachment (detach) operation in the computation of $L_{ncs}$, as proposed in the methodology.

\textbf{Sensitivity to $\beta$ and the detach operation:} Table~\ref{tab:beta} shows the model performance under different values of $\beta$, with and without the detach operation. The results indicate that, without gradient detachment (w/o detach), the model exhibits high sensitivity to $\beta$: as $\beta$ increases, $mIoU$ drops from 32.46 to 30.77. This confirms the previously mentioned risk of feature collapse, where bidirectional gradient updates can drive the model toward trivial solutions.  

In contrast, with gradient detachment (w/ detach), the model remains stable across all tested $\beta$ values and achieves its best performance at $\beta = 0.2$. This objectively demonstrates that gradient detachment is a critical mechanism for stabilizing asymmetric alignment training.

\begin{table}[htbp]
  \centering
  \caption{Comparison of the impact of the $L_{pcs}$ weight $\alpha$ on model performance. By varying $\alpha$, we seek the optimal balance for cross-modality semantic alignment on both the EarthMiss and WHU-OPT-SAR datasets.}

  \label{tab:alpha}
  \begin{tabular}{c|cc|c}
    \toprule
    $\alpha$ & EarthMiss & WHU-OPT-SAR & Mean\\
    \midrule
    0.01 & 31.21 & 32.10 & 31.66\\
    0.1  & 32.29 & 33.08 & 32.69\\
    0.2  & \cellcolor{red!10}{33.09} & 33.51 & 33.30 \\
    \cellcolor{red!10}{0.5}  & \cellcolor{red!10}{33.04} & \cellcolor{red!10}{34.07} & \cellcolor{red!25}{33.56} \\
    0.8  & 32.85 & 33.96 & 33.41 \\
    \bottomrule
  \end{tabular}
\end{table}

\textbf{Sensitivity to $\alpha$:} 
Table~\ref{tab:alpha} shows the effect of the pixel-level semantic sampling alignment loss weight $\alpha$. As $\alpha$ increases from 0.01 to 0.5, the average performance on both datasets steadily improves, peaking at $\alpha = 0.5$ with an $mIoU$ of 33.56\%. When $\alpha$ is further increased to 0.8, performance slightly declines. This indicates that a moderate level of semantic alignment constraint maximizes cross-modality feature consistency, while an excessively high weight may interfere with the model’s ability to preserve modality-specific characteristics during feature extraction.

In summary, setting $\alpha= 0.5$ strikes a balance between semantic alignment and preserving modality-specific characteristics, while $\beta = 0.2$ is sufficient to introduce asymmetry without allowing the alignment loss to dominate the optimization.

\subsubsection{Impact of Sampling Strategies}
\begin{table}[htbp]
  \centering
  \caption{Performance comparison under different pixel-level semantic sampling strategies. Here, no-sample denotes the absence of class-balanced sampling (i.e., using all pixels without balancing), and random ($N$) indicates that, within each training batch, $N$ pixel locations are randomly sampled per valid semantic class.}

  \label{tab:sampling_results}
  \begin{tabular}{l|cc|c}
    \toprule
    Sampling Method & EarthMiss & WHU-OPT-SAR & Mean \\
    \midrule
    no-sample   & 32.81 & 33.75 & 33.28 \\
    random (16) & 32.00 & 33.24 & 32.62 \\
    \rowcolor{red!10} random (32) & 33.04 & 34.07 & 33.56 \\
    random (48) & 33.01 & 33.80  & 33.40 \\
    \bottomrule
  \end{tabular}
\end{table}

To evaluate the pixel-level semantic alignment (PSA) sampling strategy described in the methodology, we compare different sampling schemes in Table~\ref{tab:sampling_results}. The results show that without any sampling (no-sample), the model fails to leverage class-balanced semantic information effectively, leading to suboptimal performance.  When random sampling is applied, the number of sampled pixels per class, denoted as $N$, significantly affects performance. At $N = 16$, the sampling density is insufficient to capture the semantic diversity of complex scenes, resulting in performance slightly worse than the no-sample baseline. Increasing $N$ to 32 yields the best $mIoU$ (33.56\%), striking an effective balance between computational efficiency and semantic representation completeness. Further increasing $N$ to 48 does not provide consistent gains, suggesting that 32 samples per class are sufficient to deliver a stable and informative supervision signal for semantic alignment under the current training framework. Accordingly, we adopt \texttt{random(32)} as the default sampling strategy in our final model.

\section{Discussion}
\label{sec:disscussion}

In the introduction, we identified two major challenges in missing-modality remote sensing semantic segmentation: (1) feature collapse and hyperparameter sensitivity in cross-modality alignment, (2) severe class imbalance causing alignment failure for minority targets. STARS addresses these interconnected problems through two synergistic core designs: the asymmetric alignment mechanism and the Pixel-level Semantic sampling Alignment (PSA) strategy.

To evaluate how these designs mitigate the identified challenges, we conduct additional analyses on the EarthMiss dataset to further validate the effectiveness of STARS from two perspectives: (1) the formation process of cross-modality semantic alignment, and (2) the disentanglement of shared and modality-specific representations. Specifically, we track the evolution of the cross-modality similarity matrix to examine how semantic consistency across modalities is gradually established for different classes, and we further analyze encoder parameter distributions and Grad-CAM~\cite{selvaraju2017grad} visualizations to understand how the shared branch learns more stable semantic cues while the modality-specific branches preserve physical details, thereby supporting robust segmentation under missing-modality settings.

\subsection{Analysis of Cross-modality Semantic Alignment}

\begin{figure}[htp]
    \centering
    \includegraphics[width=1.0\linewidth]{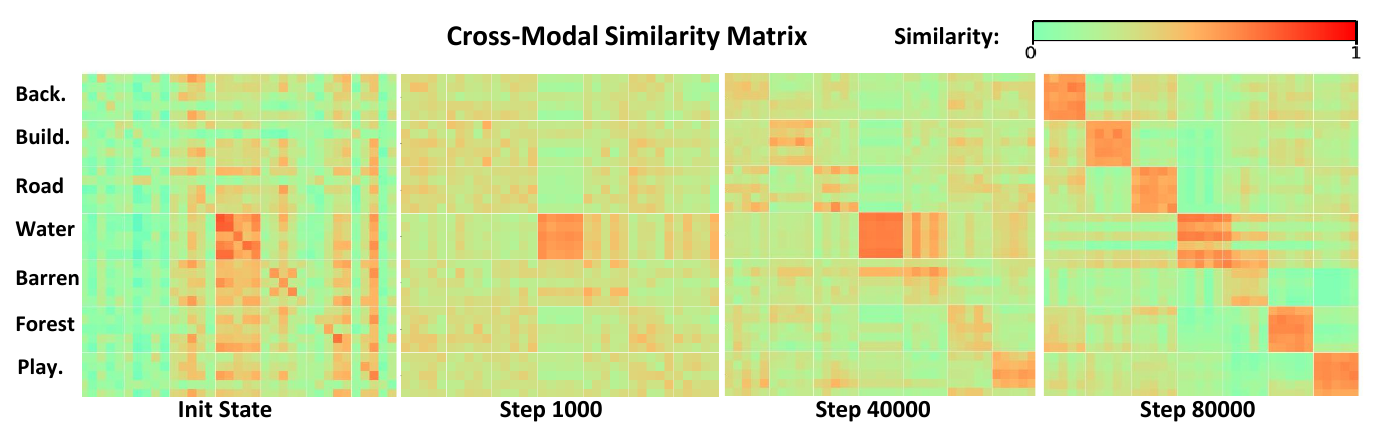}
    \caption{Evolution of the cross-modality semantic similarity matrix on  EarthMiss dataset. The rows and columns of the matrix correspond to different land cover classes, and each cell value represents the cosine similarity between feature vectors from the two modalities. As training progresses (at steps Init, 1000, 40000, and 80000), the diagonal entries become increasingly prominent, confirming the model’s ability to establish consistent cross-modality semantic associations within the shared feature space.}
    \label{fig:similarity_matrix}
\end{figure}

To monitor the formation of cross-modality semantic associations during training, we visualize the evolution of the cross-modality similarity matrix, as shown in Figure~\ref{fig:similarity_matrix}. At the initial training stage (Init State), the matrix exhibits a disordered pattern, indicating that the encoder cannot yet associate identical semantic classes across modalities. As training progresses from 1,000 to 40,000 steps, a clear diagonal structure gradually emerges. This observation directly validates the effectiveness of our proposed pixel-level semantic alignment (PSA) strategy: through class-balanced sampling and the $L_{psc}$ loss, the model successfully pulls together feature representations of the same semantic class from different modalities.

\subsection{Encoder Parameter Distribution and Modality Adaptation Analysis}

\begin{figure}[htp]
    \centering
    \includegraphics[width=1.0\linewidth]{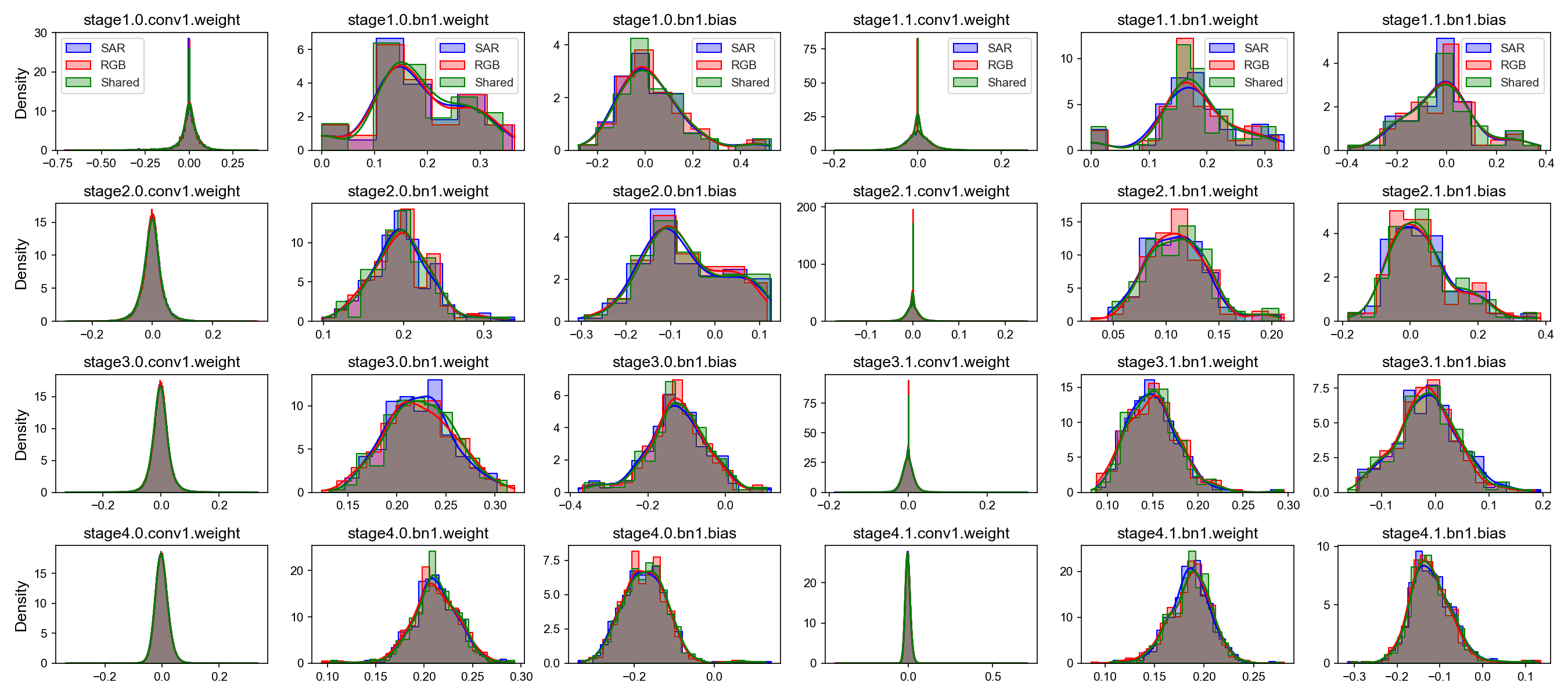}
    \caption{Comparison of parameter distributions on  EarthMiss dataset  across three encoders: SAR-specific, RGB-specific, and Shared. The figure illustrates the probability density distributions of convolutional weights (\texttt{conv1.weight}), Batch Normalization scaling factors (\texttt{bn1.weight}), and offsets (\texttt{bn1.bias}) from shallow layers (stage1) to deep layers (stage4).}
    \label{fig:distribution}
\end{figure}

To further investigate the underlying mechanism of feature disentanglement in the STARS framework, we analyze the parameter distributions of the SAR-specific encoder, RGB-specific encoder, and shared encoder, which were trained on the EarthMiss dataset, as shown in Figure~\ref{fig:distribution}.

\textbf{1. Statistical consistency of convolutional weights:}  
The convolutional weights (\texttt{conv1.weight}) of all three encoders exhibit high statistical homogeneity, following stable near-Gaussian distributions. This consistency indicates that the cross-modality alignment strategy successfully maps heterogeneous inputs onto a unified numerical manifold, effectively mitigating the modality gap. Such alignment ensures that the fusion module integrates features without bias toward any single modality due to scale discrepancies.

\textbf{2. Adaptive divergence in BN scaling factors:}  
Notably, while convolutional weights remain consistent, the \textbf{Batch Normalization (BN) scaling factors ($\gamma$)} show observable distributional differences among the three encoders—particularly in shallow layers (\textit{layer1} and \textit{layer2}). The distinct $\gamma$ distribution in the SAR encoder reflects the model’s ability to automatically suppress noise-sensitive channels and enhance physically meaningful scattering responses in SAR imagery, which differs significantly from the texture-driven activation patterns in the RGB branch. In contrast, the $\gamma$ distribution of the shared encoder is typically more concentrated, suggesting its focus on extracting cross-modality semantic primitives. These subtle parameter-level variations demonstrate that STARS does not rely on naive weight sharing; instead, it achieves genuine \textbf{semantic disentanglement} through channel-wise recalibration in the BN layers.

\textbf{3. Layer-wise evolution:}  
The distributional differences are most pronounced in lower layers and gradually converge with increasing depth (e.g., in \textit{layer4}). This aligns with the hierarchical nature of deep networks: shallow layers capture modality-specific low-level features (e.g., speckle noise in SAR and color gradients in RGB), while deeper layers progressively evolve toward abstract, modality-invariant semantic representations. This statistical stability ensures that, during inference under optical-modality absence, the SAR-only branch can still preserve the semantic discriminative logic learned during joint training.

This distribution pattern provides empirical evidence that STARS successfully circumvents the feature collapse issue highlighted in Section~\ref{sec:intro}. In trivial solutions often plagued by symmetric alignment, the encoders tend to output constant vectors, leading to homogenized parameter distributions. However, Figure~\ref{fig:distribution} shows that while the convolutional layers maintain a shared semantic manifold (Gaussian consistency), the BN layers actively adapt to modality-specific statistics (Distinct distributions). This confirms that our asymmetric gradient control effectively forces the model to learn meaningful, modality-specific transformations rather than collapsing into a degenerate identity mapping, thereby solving the instability issue found in traditional Siamese-based alignment methods.

\subsection{Class Activation Mapping Analysis}

\begin{figure}[htp]
    \centering
    \includegraphics[width=1.0\linewidth]{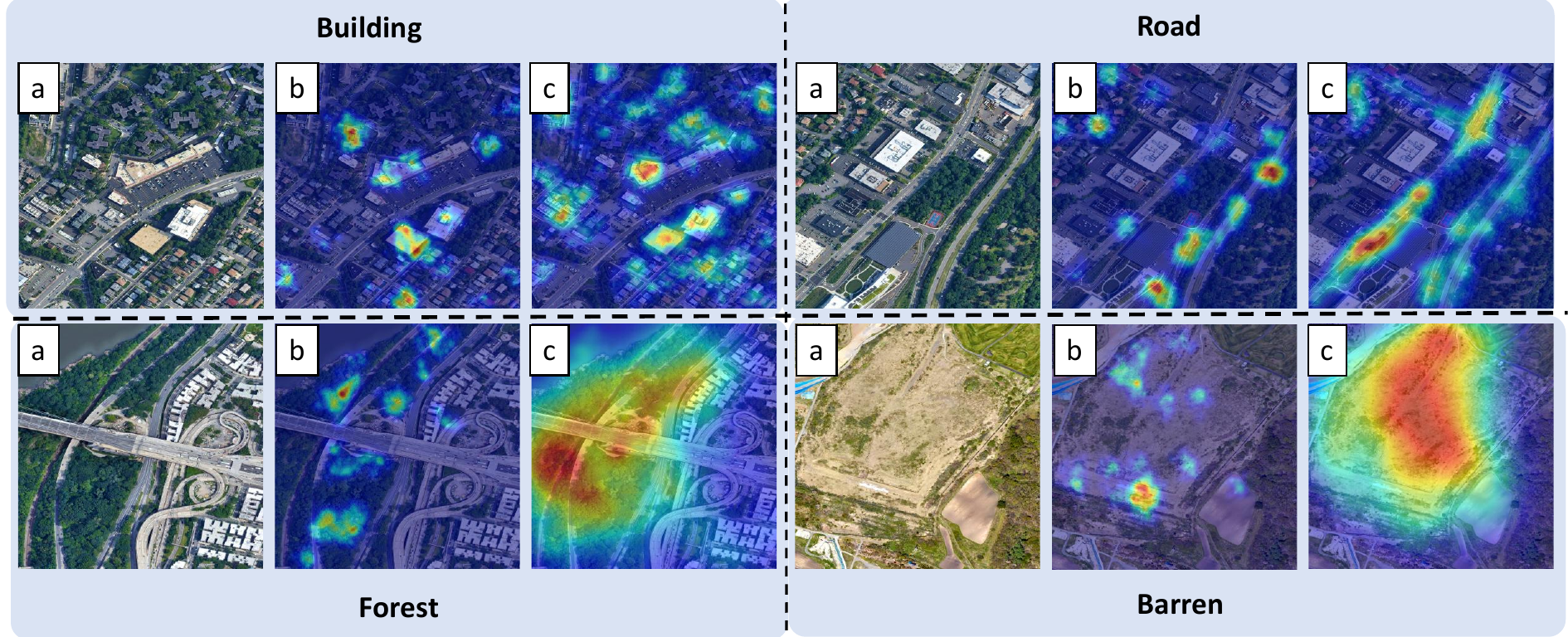}
    \caption{Comparison of Grad-CAM visualizations from different encoder branches on  EarthMiss dataset. (a) Original image (reference); (b) activation map from the SAR-specific encoder, showing sensitivity to physical scattering points; (c) activation map from the shared encoder, demonstrating stronger semantic continuity and an improved capability in capturing spatial object boundaries.}
    \label{fig:gradcam}
\end{figure}

Using Grad-CAM~\cite{selvaraju2017grad} to visualize and compare activation regions of different encoder branches (see Figure~\ref{fig:gradcam}), we can directly observe the feature extraction characteristics of the STARS framework. The results show that the SAR-specific encoder is highly sensitive to physical scattering properties, with activations appearing as scattered or fragmented spots primarily concentrated along strong-scattering geometric edges. This reflects the intrinsic physical nature of SAR imagery but reveals a limited ability to perceive complete semantic object boundaries.

In contrast, the shared encoder demonstrates superior semantic aggregation capability. Its activation map fully covers building regions and maintains linear continuity for road structures, exhibiting spatial coherence that closely aligns with human visual perception from optical imagery. These visualizations confirm that, even in the absence of optical input during testing, the shared encoder effectively guides the model to accurately localize semantic regions, thereby compensating for the inherent semantic deficiencies of SAR imagery.

Crucially, the clear activation boundaries observed in the shared encoder (Figure~\ref{fig:gradcam}(c)) directly validate the efficacy of our Pixel-level Semantic Sampling Alignment (PSA) strategy in addressing the class imbalance problem. As discussed in Section~\ref{sec:intro}, standard global alignment tends to be dominated by background classes, causing the model to miss the structural details of minority targets like buildings. By enforcing balanced semantic constraints, PSA compels the network to attend to the complete spatial extent of these targets. The contrast between the fragmented activations in the SAR branch and the coherent activations in the Shared branch confirms that STARS has successfully transferred the structural priors from the optical modality to the shared representation, mitigating the marginalization of minority key targets typically seen in previous works.

\section{Conclusion}
\label{sec:conclusion}

Multimodal remote sensing semantic segmentation holds great promise for applications such as land cover mapping and urban planning. However, the frequent absence of auxiliary modalities in real-world scenarios severely limits its practical deployment. Conventional fusion models are fragile under modality-missing conditions and suffer from issues including feature collapse, alignment failure due to class imbalance, and insufficient generalization of recovered features.

We propose a novel and robust semantic segmentation framework, STARS (Shared-specific Translation and Alignment for missing-modality Remote Sensing), to address the challenge of semantic segmentation in multimodal remote sensing imagery under modality-missing conditions. We introduce an asymmetric alignment mechanism based on bidirectional translation and gradient stop, which reduces the model’s sensitivity to  loss weights. Additionally, we design a pixel-level semantic sampling alignment strategy that explicitly pulls together semantically consistent cross-modality pixel pairs and pushes apart dissimilar ones in the feature space, thereby improving the recognition accuracy of minority-class targets.

Extensive experiments on multiple multimodal remote sensing datasets—including EarthMiss, WHU-OPT-SAR, and ISPRS Potsdam—demonstrate the effectiveness of STARS. Under modality-missing test conditions, STARS achieves higher $mIoU$ and $mF1$ scores than state-of-the-art methods. Ablation studies confirm the effectiveness of core components, including asymmetric alignment, pixel-level semantic sampling alignment, and bidirectional translation. Furthermore, analyses of cross-modality semantic similarity matrices, encoder parameter distributions, and Grad-CAM visualizations provide deep insights into how STARS establishes robust cross-modality semantic associations, disentangles modality-specific features, and maintains both stability and interpretability in feature representation.

The STARS framework offers an effective and reliable solution to the modality-missing problem in remote sensing. Future work will explore extending STARS to multimodal scenarios (with more than two modalities) and to settings where arbitrary combinations of modalities may be missing, and extend STARS to other multimodal remote sensing applications such as object detection and change detection.

\section*{Acknowledgement}

The numerical calculations in this paper have been done on the supercomputing system in the Supercomputing Center of Wuhan University. Tribute to open-source contributors, and gratitude to the authors of open-source works such as PyTorch, GDAL, Grad-CAM.

\section*{Declaration of generative AI and AI-assisted technologies in the manuscript preparation process.}
During the preparation of this work the author(s) used GPT-4o in order to grammatical modification and polish. After using this tool/service, the author(s) reviewed and edited the content as needed and take(s) full responsibility for the content of the published article.

\section*{Funding}
This work was supported by the National Natural Science Foundation of China [grant number 42101346]; the Science and Technology Commission of Shanghai Municipality [grant number 22DZ1100800]; and the China Postdoctoral Science Foundation [grant number 2020M680109].

\printcredits

\bibliographystyle{elsarticle-num-names} 
\bibliography{main}

\end{document}